\documentclass[10pt, journal, compsoc]{IEEEtran}
\usepackage{array,multirow}
\usepackage{tabu}
\usepackage{subcaption}
\usepackage{graphicx}
\usepackage{amsmath}

\begin{document}

\title{Face Retrieval using Frequency Decoded Local Descriptor}

\author{
Shiv Ram Dubey
\thanks{
\vspace{0.5cm}

S.R. Dubey is associated with the Computer Vision Group at Indian Institute of Information Technology, Sri City, Andhra Pradesh - 517646, India
\vspace{0.3cm}

Preprint: Accepted in Multimedia Tools and Applications, Springer}
}

\markboth{Preprint: Accepted in Multimedia Tools and Applications, Springer}
{Shell \MakeLowercase{\textit{et al.}}: Bare Demo of IEEEtran.cls for IEEE Journals}

\IEEEtitleabstractindextext{%
\begin{abstract}
The local descriptors have been the backbone of most of the computer vision problems. Most of the existing local descriptors are generated over the raw input images. In order to increase the discriminative power of the local descriptors, some researchers converted the raw image into multiple images with the help of some high and low pass frequency filters, then the local descriptors are computed over each filtered image and finally concatenated into a single descriptor. By doing so, these approaches do not utilize the inter frequency relationship which causes the less improvement in the discriminative power of the descriptor that could be achieved. In this paper, this problem is solved by utilizing the decoder concept of multi-channel decoded local binary pattern over the multi-frequency patterns. A frequency decoded local binary pattern (FDLBP) is proposed with two decoders. Each decoder works with one low frequency pattern and two high frequency patterns. Finally, the descriptors from both decoders are concatenated to form the single descriptor. The face retrieval experiments are conducted over four benchmarks and challenging databases such as PaSC, LFW, PubFig, and ESSEX. The experimental results confirm the superiority of the FDLBP descriptor as compared to the state-of-the-art descriptors such as LBP, SOBEL\_LBP, BoF\_LBP, SVD\_S\_LBP, mdLBP, etc.
\end{abstract}

}

\maketitle


\section{Introduction}
\subsection{Motivation}
\IEEEPARstart{F}{acial} analysis is a very active research area in the computer vision field. In the early days, the face recognition was being done in a very controlled environment, where the images were being captured in frontal pose with uniform background, etc. From last decade, the research focus has been shifted towards the unconstrained and robust face recognition. Wright et al. followed a sparse representation for face recognition \cite{wright2009}. Ding et al. did the pose normalization for face recognition against pose variations \cite{ding2012}. The pose-invariant face recognition is converted into a partial frontal face recognition and then a patch-based face description is employed in \cite{ding2015multi}. Face recognition against pose, illumination and blur is conducted by Punnappurath et al. by modeling the blurred face as a convex set \cite{abhi2015}. The survey in face recognition is also conducted by several researchers \cite{zhao2003face}, \cite{zhang2009face}, \cite{ding2016comprehensive}. Due to the increasing demand of real life facial analysis, unconstrained and robust algorithms need to be investigated.

\subsection{Related Works}
The facial analysis approaches can be categorized into the following three areas: deep learning based, learning descriptor based, and handcrafted based. The deep learning based approaches are proposed recently such as DeepFace \cite{deepface} and FaceNet \cite{facenet}. These approaches have gained very appealing result, but at the cost of huge computation. More complex computing resources as well as huge amount of data are required to run the deep learning based approaches. Moreover, the database biased-ness has been seen in the deep learning based approaches. Several learning based descriptors have been investigated for the face recognition task, such as learning based descriptor (LE) \cite{cao2010}, descriptor with learned background \cite{wolf2011}, learned discriminant descriptor \cite{lei2014}, learned binary descriptor \cite{lu2015learning}, and simultaneous binary learning and encoding \cite{lu2015simultaneous}. These descriptors rely over some codebook or clustering methods to learn the training data statistics and cannot learn the features that are not present in the training data. Whereas, on the other hand, the handcrafted descriptors are easy to design and follow, independent of the data, do not require complex computing resources and huge amount of data for the training. In the rest of this paper, the hand-designed descriptors are the focus of interest.

The local binary pattern (LBP) is a widely adapted local descriptor \cite{lbptexture}. It encodes the local relationship of the pixels. Basically, it finds a binary pattern for each pixel and represent it into the histogram form. A binary pattern for a pixel consists of the 8 binary values corresponding to 8 local neighbors of the concerned pixel distributed in a circular fashion. For a neighbor, the binary value is 1 if the intensity value of neighboring pixel is greater than or equal to the intensity value of the center pixel, otherwise the binary value is 0. Originally, LBP was proposed for the texture classification \cite{lbptexture}. However, later on, it is proved as the foundation to solve many computer vision problems \cite{lbpbook}. Several variants of LBP have been investigated by several researchers for various problems such as local image matching \cite{iold}, content based image retrieval \cite{ltrp}, \cite{boflbp}, \cite{mdlbp}, texture classification \cite{dlbp}, \cite{brint}, \cite{priclbp}, biomedical image analysis \cite{ldep}, \cite{lbdp}, \cite{lwp}, \cite{lbdisp}, etc.

In 2006, Ahonen et al. utilized the LBP concept for face representation and recognition which is later regarded as the state-of-the-art in face recognition \cite{lbp}. Inspired from the success of LBP in face recognition as well as other problems, many improvements in LBP have been proposed for the face recognition \cite{lbpbook}. Huang et al. presented a literature survey over LBP based descriptors in 2011 for the facial image analysis \cite{huang2011local}. In 2013, one more survey over LBP based face recognition is conducted by Yang and Chen \cite{yang2013comparative}. A local ternary pattern (LTP) is proposed in \cite{ltp} for illumination robust face recognition by extending the binary concept of LBP into ternary with the help of a threshold. Finding the suitable threshold in LTP is one of the problems. The local derivative pattern (LDP) utilizes the LBP concept over high order derivatives \cite{ldp}. Basically, LDP first computes the four derivative images in $0^o$, $45^o$, $90^o$, and $135^o$ directions respectively. Then, it finds the LBP histogram over each derivative image and finally concatenate into a single descriptor. Vu et al. proposed the patterns of oriented edge magnitudes (POEM) descriptor by incorporating the gradient magnitude and orientation information \cite{poem}. In order to further improve the POEM descriptor, the whitened principal-component-analysis dimensionality reduction technique is adapted by Vu and Caplier \cite{vu2012}. Later on, in 2013, Vu proposed patterns of orientation difference (POD) descriptor and combined with the POEM descriptor to boost the performance for face recognition \cite{vu2013}.

A local vector pattern (LVP) is investigated by Fan and Hung from the vector representations of each pixel \cite{lvp}. The vector representation of a center pixel in LVP utilizes the relationship between the center pixel and its adjacent neighbors at various distances in different directions. In dimensionality reduced local directional pattern (DR-LDP) \cite{drldp}, first a local directional pattern is computed for each pixel of the face image, then the image is divided into multiple blocks, and finally a single code per block is generated by XORing the local directional pattern codes of each pixel of that block. The local directional pattern uses the low-pass Kirsch filters in eight directions to get the directional information. Ding et al. introduced multi-directional multi-level dual-cross pattern (MDML-DCP) for face images \cite{dcp}. In order to reduce the impact of differences in illumination, they employed the first derivative of a Gaussian with varying scale. The dual-cross pattern is then computed from the multi-radius neighborhood in two different neighboring sets, one consisting of horizontal and vertical neighbors and another one consisting of diagonal neighbors. The gradient orientations in two directional gradient responses in horizontal and vertical directions are used to design the local gradient order pattern (LGOP) for face representation \cite{lgop}. Recently, a local directional gradient pattern (LDGP) exploited the relationship between the referenced pixel in four different directions in high order derivative space for face recognition \cite{ldgp}. The local directional ternary pattern (LDTP) uses eight Robinson compass masks as the low-pass filters to generate the eight directional images and then computes the final ternary pattern based on the primary and secondary directions \cite{ldtp}.

Most of the LBP variants have focused over the derivative, direction and coding pattern and overlooked the frequency information. Some descriptors used high-pass information, but only for the direction purpose. The high-pass information in the image corresponds to the details and low-pass information corresponds to the coarse of the image. Zhao et al. \cite{sobel-lbp} proposed the Sobel local binary pattern (SOBEL\_LBP) from the Sobel horizontal and vertical filters for face representation. In SOBEL\_LBP, two filtered images are obtained using Sobel operators, then the LBP histograms are computed over both and concatenated to produce the final descriptor. In order to increase the robustness of the descriptor, a pre-processing in the form of high-pass frequency filter is used in semi-structure local binary pattern (SLBP) for facial analysis \cite{slbp}. The $3\times3$ mask of filter used in SLBP is [1,1,1;1,1,1;1,1,1]. 
Recently, a bag-of-filtered local binary pattern (BoF\_LBP) is proposed for content-based image retrieval \cite{boflbp}. BoF\_LBP concatenates the LBP computed over the several filtered images. Five filters, including one low-pass and four high-pass filters are used in BoF\_LBP.
The local SVD decomposition has recently been used as the pre-processing step with the existing descriptors such LBP to develop the SVD\_S\_LBP descriptor \cite{svdslbp}. It is explored in \cite{svdslbp} that the S sub-band of SVD is more discriminative and worth for the face retrieval over near-infrared images.
A concept of multi-channel decoded local binary pattern (mdLBP) is proposed recently for color image retrieval, where the LBP is first computed over each color channel separately and then passed through a decoder to utilize the inter-channel relationship in the final feature vector \cite{mdlbp}. Some recent color descriptors are local color occurrence descriptor \cite{lcod}, rotation invariant structure based descriptor \cite{rshd}, illumination invariant color descriptor \cite{ic}, etc. Other notable image retrieval methods are based on SURF binarization and fast codebook construction \cite{kan2017surf}, feature fusion using compressed sensing \cite{wang2018compressed}, and feature fusion using separable vocabulary \cite{wang2017separable}.

\subsection{Contributions}
In this paper, the decoder concept is utilized with the multi-frequency filtered images. Instead of simply concatenating the LBP features computed over different filtered image in BoF\_LBP or SOBEL\_LBP, here it is passed to the decoder to encode the inter-frequency information and then the output channels are combined. The BoF\_LBP has used five filters with one low-pass and four high-pass. The dimension of the final descriptor will be quite high if all five frequencies are used with one decoder. In order to resolve this problem, two decoders with one low-pass frequency information and two high-pass frequency information are used in this work. The combination of different frequency information is evaluated empirically. The main contributions of this paper are summarized as follows:
\begin{itemize}
\item Multiple frequency information, including one low-pass and four high-pass is used to increase the discriminative power and robustness of the descriptor.
\item In the proposed frequency decoded local binary pattern (FDLBP), the inter-frequency information is encoded with the help of decoder concept using LBP binary bits.
\item In order to reduce the dimensionality of the descriptor, two decoders are used with one low-pass and two high-pass filters. Different frequency decoder combinations are also explored through the experiment.
\item The concept of the proposed frequency decoder is also extended for the color face images.
\item The image retrieval experiments are conducted over four very challenging face databases including Point and Shoot Challenge, Labeled Faces in Wild, Public Figures, and ESSEX.
\item The proposed descriptor is tested with different distance measures such as Euclidean, Cosine, Emd, L1, D1, and Chi-square.
\end{itemize}

The rest of the paper is structured in the following manner: Section II describes the proposed descriptor; Section III illustrates the experimental setup; Section IV reports the experimental results, comparison and analysis; and finally Section V sets the concluding remarks.
\begin{figure*}[!t]  
  \centering
  \begin{subfigure}{.33\textwidth}
    \centering
    \includegraphics[width=0.68\linewidth]{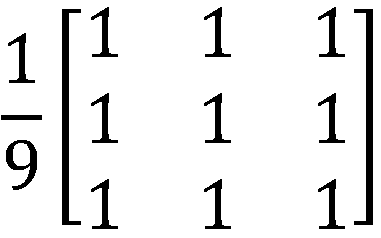}
    \caption{$F_a$ Filter}
    \label{fa}
  \end{subfigure}%
  \begin{subfigure}{.33\textwidth}
    \centering
    \includegraphics[width=0.85\linewidth]{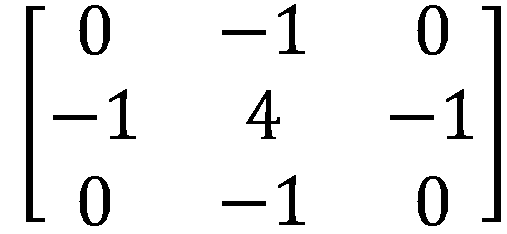}
    \caption{$F_{hv}$ Filter}
    \label{fhv}
  \end{subfigure}%
  \begin{subfigure}{.33\textwidth}
    \centering
    \includegraphics[width=0.8\linewidth]{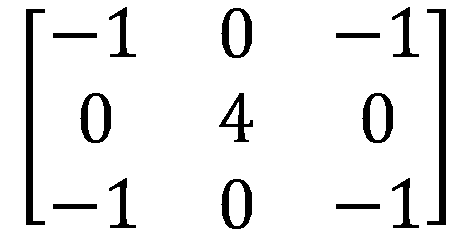}
    \caption{$F_d$ Filter}
    \label{fd}
  \end{subfigure}
  \begin{subfigure}{.33\textwidth}
    \centering
    \includegraphics[width=0.85\linewidth]{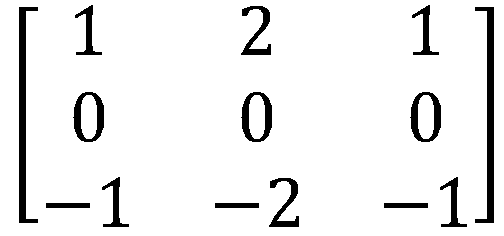}
    \caption{$F_{SV}$ Filter}
    \label{fsv}
  \end{subfigure}%
  \begin{subfigure}{.33\textwidth}
    \centering
    \includegraphics[width=.68\linewidth]{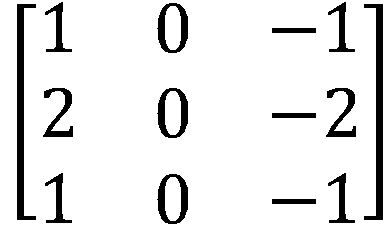}
    \caption{$F_{SH}$ Filter}
    \label{fsh}
  \end{subfigure}
  \vspace{2mm}
  
  \caption{The five filters, including one low-pass and four high-pass filters used in this paper. The $F_a$, $F_{hv}$, $F_d$, $F_{SV}$, and $F_{SH}$, denotes the Average Filter, Horizontal-vertical Filter, Diagonal Filter, Sobel Vertical Edge Filter, and Sobel Horizontal Edge Filter, respectively.}
  \label{filters}
\end{figure*}

\begin{figure*}[!t]  
  \centering
  \begin{subfigure}{.16\textwidth}
    \centering
    \includegraphics[width=.98\linewidth]{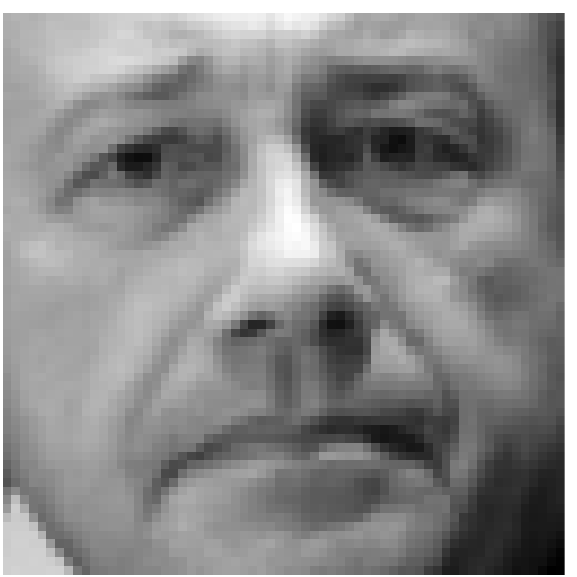}
    \caption{$I$}
    \label{original}
  \end{subfigure}
  \hspace{0.21cm}
  \begin{subfigure}{.16\textwidth}
    \centering
    \includegraphics[width=.98\linewidth]{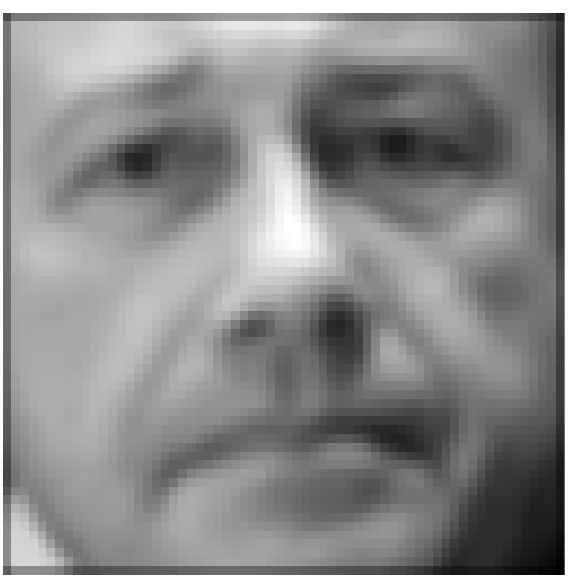}
    \caption{$F_{a}$}
    \label{fa}
  \end{subfigure}%
  \begin{subfigure}{.16\textwidth}
    \centering
    \includegraphics[width=.98\linewidth]{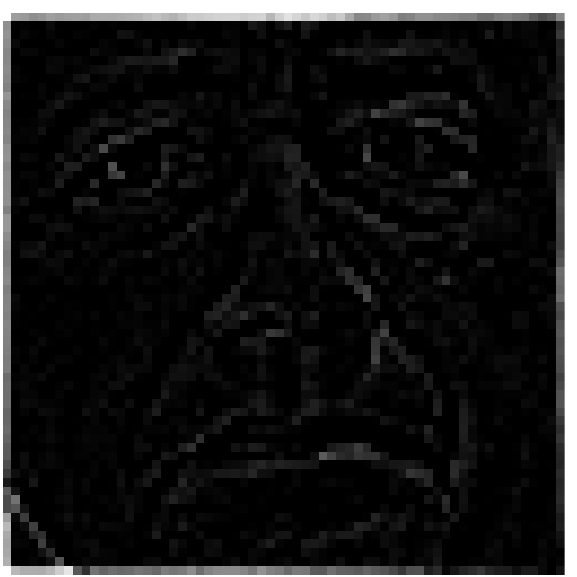}
    \caption{$F_{hv}$}
    \label{fhv}
  \end{subfigure}%
  \begin{subfigure}{.16\textwidth}
    \centering
    \includegraphics[width=.98\linewidth]{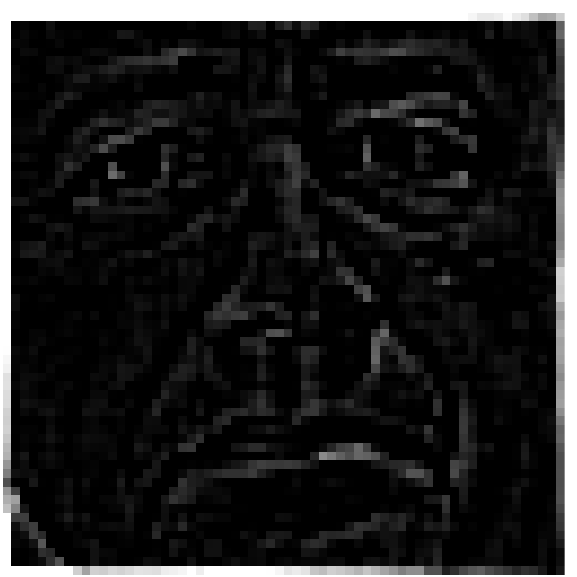}
    \caption{$F_d$}
    \label{fd}
  \end{subfigure}%
  \begin{subfigure}{.16\textwidth}
    \centering
    \includegraphics[width=.98\linewidth]{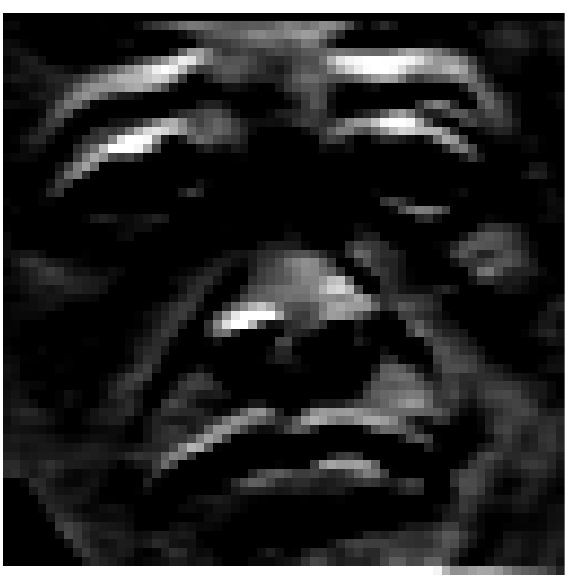}
    \caption{$F_{SV}$}
    \label{fsv}
  \end{subfigure}
  \begin{subfigure}{.16\textwidth}
    \centering
    \includegraphics[width=.98\linewidth]{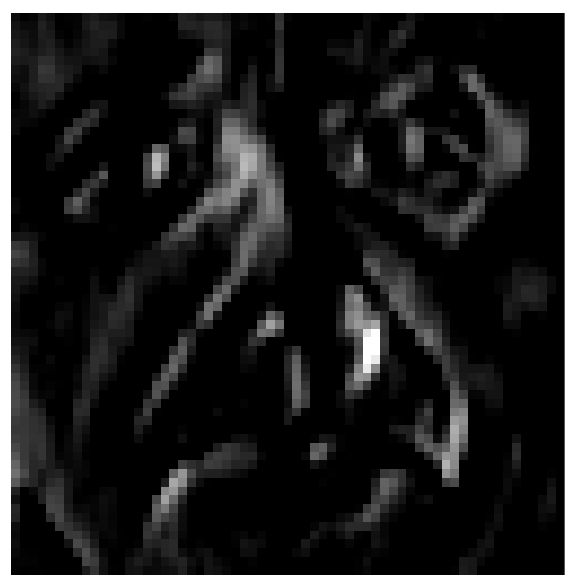}
    \caption{$F_{SH}$}
    \label{fsh}
  \end{subfigure}

  \vspace{2mm}  
  \caption{An example demonstrating the filters used to generate the difference frequency images. (a) Original Image, $I$, taken from the LFW database \cite{lfw}. Images obtained using (b) $F_a$, i.e., the Average Filter, (c) $F_{hv}$, i.e., the Horizontal-vertical Filter, (d) $F_d$, i.e., the Diagonal Filter, (e) $F_{SV}$, i.e., the Sobel Vertical Edge Filter, and (f) $F_{SH}$, i.e., the Sobel Horizontal Edge Filter.}
  \label{filter:example}
\end{figure*}

\section{The Proposed FDLBP Descriptor}
In this section, the proposed frequency decoded local binary pattern (FDLBP) computation process is described in detail. The whole procedure is divided into several steps such as image filtering with low-pass and high-pass filters, LBP binary code computation, inter-frequency coding, feature vector computation, and color channel consideration.

\subsection{Image Filtering}
The low-pass filters generate the coarse information, whereas the high-pass filters generate the detailed information of the image. The image filtering with one low-pass and four high-pass filters are done in this paper. These filters are the same as used in BoF\_LBP \cite{boflbp} and shown in  Fig. \ref{filters}. The average filter denoted by $F_a$ is used for the low-pass frequency filtering. The horizontal-vertical difference filter (i.e., $F_{hv}$), diagonal difference filter (i.e., $F_{d}$), Sobel vertical edge filter (i.e., $F_{sv}$), and Sobel horizontal edge filter (i.e., $F_{hh}$) are used for the high-pass frequency filtering. Suppose, the input grayscale image is represented by $I$ with dimension $x \times y$ (i.e. image $I$ is having $x$ rows and $y$ columns) and the filtered images are denoted by $I_{a}$, $I_{hv}$, $I_{d}$, $I_{sv}$, and $I_{sh}$ by using $F_{a}$, $F_{hv}$, $F_{d}$, $F_{sv}$, and $F_{sh}$ filters respectively. Mathematically, $I_k|_{\forall k=a,hv,d,sv,sh}$ can be given as follows,
\begin{equation}
I_k=I*F_k
\end{equation}
where '$*$' represents the 2-D convolution operation between two matrices and $F_k$ is a $3\times3$ filter with $k \in [a,hv,d,sv,sh]$. Note that the input image $I$ is padded with one pixel (i.e., nearly half of the size of the filter) in all directions to obtain the same size filtered image $I_k$. 

The Fig. \ref{filter:example} presents the five filtered face images for an input face image using the five filters, $F_k|_k\in[a,hv,d,sv,sh]$. The input face image is taken from the LFW database \cite{lfw}. It can be seen that the filtered image using a low-pass filter (i.e., $F_a$) in Fig. \ref{original} is having the coarse information of the face image, whereas other hand the filtered images using high-pass filters (i.e., $F_{hv}$, $F_d$, $F_{sv}$, and $F_{sh}$) are having the detailed information with corresponding edges. More specifically, Fig. \ref{fhv} and \ref{fd} is having the horizontal-vertical difference  and diagonal difference information, whereas, Fig. \ref{fsv} and \ref{fsh} are having the horizontal edge and vertical edge information.

\subsection{LBP Binary Code Generation}
\begin{figure}[!t]
    \centering
    \includegraphics[width=.8\linewidth]{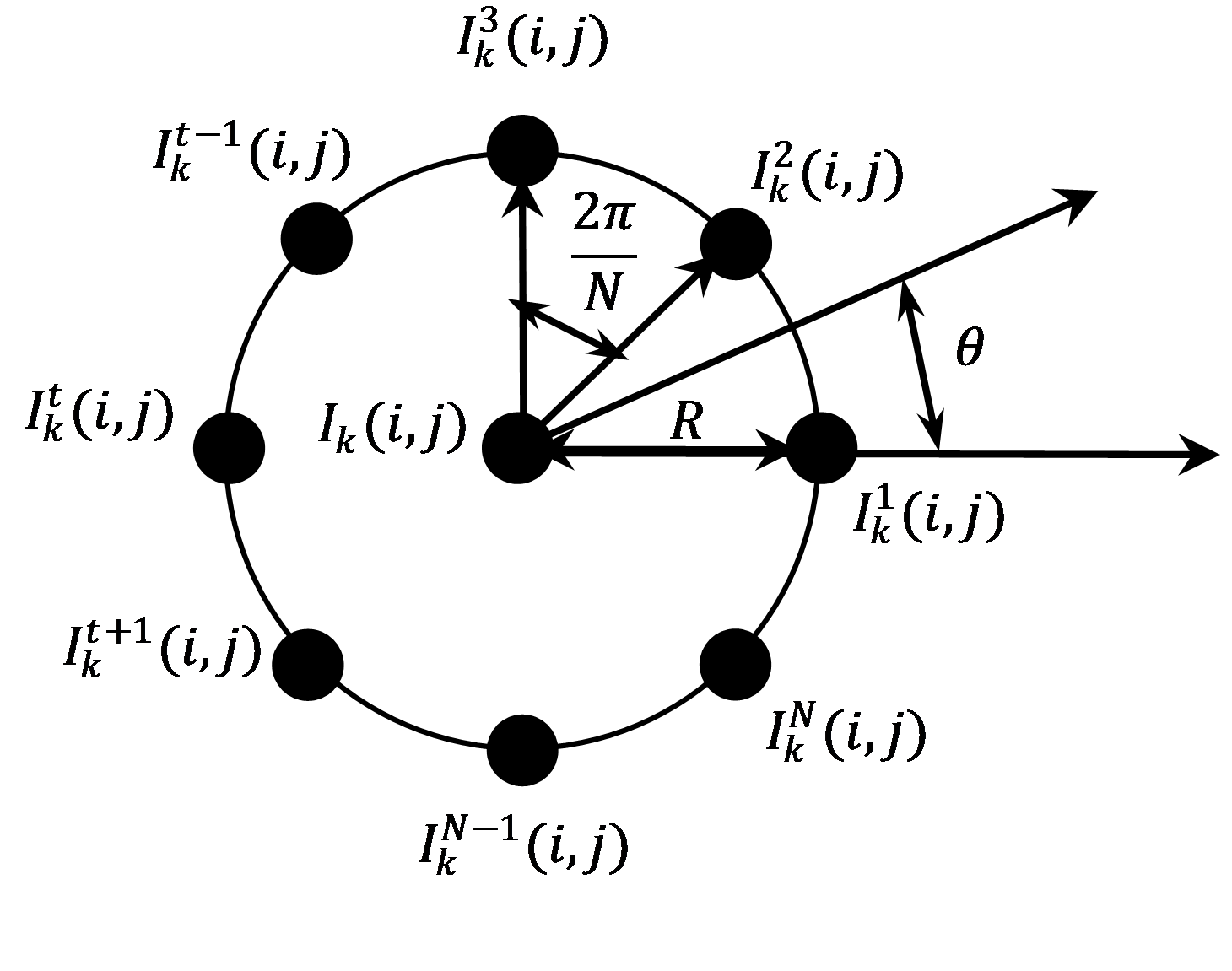}
    \caption{The $N$ neighbors $I^t_k(i,j)|_{\forall t \in [1,N]}$ in the local neighborhood of pixel $I_k(i,j)$ at radius $R$, where $k \in [a,hv,d,sv,sh]$. Note that the first neighbor is considered in the right side and rest neighbors are considered with regard to the first neighbor in counter-clockwise direction.}
    \label{neighbors}
\end{figure}

Once the input face image is transferred into filtered domain using different filters, the local binary pattern (LBP) \cite{lbp} codes are generated for each pixel of each filtered face image. As per the convention, $I_k$ for $k \in [hv, d, sv, sh]$ is the input to the LBP code generator. Suppose $I_k(i,j)$ is the value in $i^{th}$ row and $j^{th}$ column of filtered image $I_k$. Consider, LBP is generated from the $N$ number of local neighbors evenly distributed in a circular fashion at a radius $R$. The number of binary bits in the LBP code is same as the number of local neighbors, i.e., $N$. The $t^{th}$ neighbor for a reference pixel $I_k(i,j)$ is denoted by $I^t_k(i,j)$, where $t \in [1,N]$ as shown in Fig. \ref{neighbors}. The $1^{st}$ neighbor is considered in the right side of the reference pixel and rest neighbors are considered with regard to the $1^{st}$ neighbor in a circular fashion in the counter-clockwise direction (see Fig. \ref{neighbors}). The coordinate of $t^{th}$ neighbor of center pixel $I_k(i,j)$ can be given by $(i-R \times sin(\theta_t)$, $j-R \times cos(\theta_t))$, where $\theta_t$ is the angular displacement of $t^{th}$ neighbor with regard to the $1^{st}$ neighbor and computed as follows,
\begin{equation}
\theta_t = (t-1) \times \theta
\end{equation}
where $\theta$ is defined as the angular displacement between any two adjacent neighbors and can be written as follows,
\begin{equation}
\theta = \frac{360}{N}.
\end{equation}

The $N$ bit binary LBP code for a reference pixel $I_k(i,j)$ is computed as follows,
\begin{equation}
\beta_k(i,j) = [\beta^1_k(i,j), \beta^2_k(i,j), ...,\beta^t_k(i,j),...,\beta^N_k(i,j)]
\end{equation}
where $\beta^t_k(i,j)$ is the binary bit value computed for the $t^{th}$ neighbor of reference pixel $I_k(i,j)$. The $\beta^t_k(i,j)$ is generated from the center value $I_k(i,j)$ and $t^{th}$ neighboring value $I^t_k(i,j)$ as follows,
\begin{equation}
\beta^t_k(i,j) = sign(\Delta(I^t_k(i,j), I_k(i,j)))
\end{equation}
where $\Delta(\alpha_1, \alpha_2)$ encodes the amount of change in between $\alpha_1$ and $\alpha_2$ and derived as follows,
\begin{equation}
\Delta(\alpha_1, \alpha_2) = \alpha_1 - \alpha_2
\end{equation}
and $sign(\alpha)$ converts the $\alpha$ in binary form with the help of following rule,
\begin{equation}
sign(\alpha) = 
\begin{cases}
1,		& \text{if } \alpha \geq 0\\
0,      & \text{otherwise}
\end{cases}
\end{equation}
In the next sub-section, the LBP binary bits (i.e., $\beta_k$) computed over each filtered image (i.e., $\forall k \in [a, hv, d, sv, sh]$) are used as the input to the frequency decoder.

\begin{figure*}[!t]
    \centering
    \includegraphics[width=.99\linewidth]{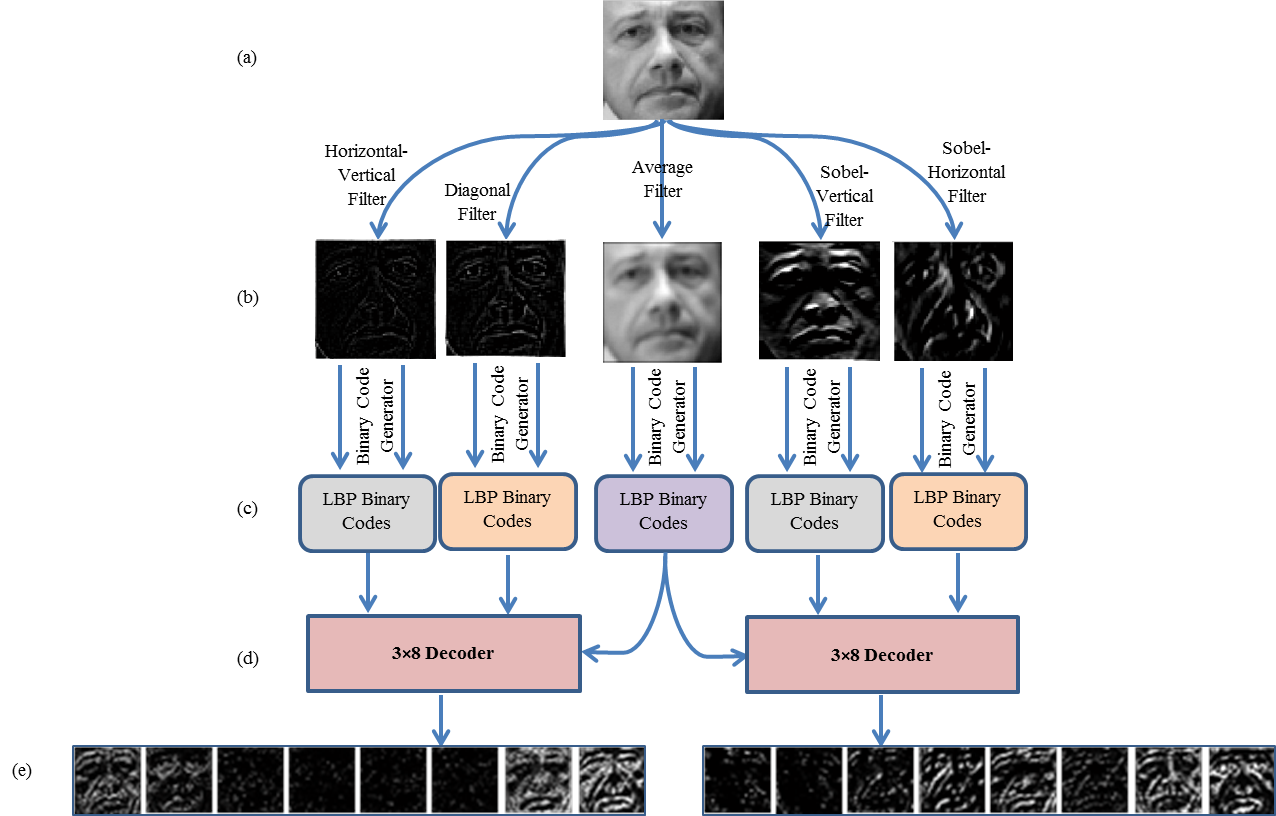}
    \caption{The illustration of frequency decoded local binary pattern. (a) An example face image taken from LFW database \cite{lfw}. (b) Five images obtained by applying the four high-pass filter and one low-pass filter (i.e., middle one). These filters are also used in BoF\_LBP \cite{boflbp}. (c) The LBP binary $N-bit$ code computed for every pixel of filtered face images using LBP operator \cite{lbp}. (d) The concept of decoder is used to utilize the inter-frequency relationship of LBP binary representation. Note that the decoder concept in mdLBP is applied over the Red, Green and Blue color channels of the image \cite{mdlbp}, whereas it is applied over the different filtered images, including one low-pass filter and two high-pass filters in this work. (e) The sixteen output face representations, including eight from each decoder are shown for the considered example face image.}
    \label{decoder-example}
\end{figure*}

\subsection{Frequency Decoder}
The concept of decoder is used in this work to utilize the relationship among different frequency filtered images. The decoder concept is also used in mdLBP with the Red, Green, and Blue color channels of the image \cite{mdlbp}. Here, in this work, two decoders are used to control the number of output images. The number of output channels for a single decoder is $2^\gamma$, where $\gamma$ is the number of input channel. So, in case of one decoder, the number of output image will be $2^5=32$ for $\gamma = 5$ input images. Whereas, two decoders with $\gamma = 3$ input image each, have the $2 \times 2^3 = 16$ output images. So, in multi-decoder framework with a $\Gamma$ number of decoders, each having a $\gamma$ number of input images, the total number of output images is $\Gamma \times 2^\gamma$. Suppose, $\Psi$ represents the output image of the decoders. The $\zeta^{th}$ output of the $\eta^{th}$ decoder is given by $\Psi^\zeta_\eta$, where $\zeta \in [1, 2^\gamma]$ and $\eta \in [1, \Gamma]$. The $\Psi^\zeta_\eta (i,j)$ is the value of face representation in the $i^{th}$ row and $j^{th}$ column of $\zeta^{th}$ output image of the $\eta^{th}$ decoder and given as follows,
\begin{equation}
\Psi^\zeta_\eta (i,j) = \sum_{t=1}^{N} \Psi^{\zeta, t}_\eta (i,j) \times \omega_t
\label{eq:output}
\end{equation}
where $\omega_t$ is a weighting factor for $t^{th}$ bit and computed as follows,
\begin{equation}
\omega_t|_{t \in [1, N]} = (2)^{(t-1)}
\end{equation}
and $\Psi^{\zeta, t}_\eta (i,j)$ is the $t^{th}$ binary value for pixel $(i, j)$ in the $\zeta^{th}$ output image of $\eta^{th}$ decoder and calculated with the help of input binary bits.

In this paper, $\Gamma=2$ number of decoders are used and $\gamma=3$ number of inputs are given to each decoder. So, each decoder generates $\zeta=2^\gamma=8$ number of output images. One of the inputs to both decoders is the LBP binary codes of low-pass filtered image with average filtering, i.e., $\beta_a$. The rest of two inputs to $1^{st}$ decoder are the LBP binary codes of high-pass filtered images with horizontal-vertical and diagonal filtering, i.e., $\beta_{hv}$, and $\beta_{d}$ respectively. Similarly, the other two inputs to $2^{nd}$ decoder are the LBP binary codes of high-pass filtered images with Sobel-vertical and Sobel-horizontal filtering, i.e., $\beta_{sv}$, and $\beta_{sh}$ respectively. So, as per the above convention, the output of $1^{st}$ and $2^{nd}$ decoder is denoted by $\Psi^{\zeta}_1$ and $\Psi^{\zeta}_2$ respectively. The $\Psi^{\zeta, t}_1 (i,j)$ is the $t^{th}$ binary value for pixel $(i, j)$ in the $\zeta^{th}$ output image of $1^{st}$ decoder and calculated as follows,
\begin{equation}
\Psi^{\zeta, t}_1 (i,j) = 
\begin{cases}
1,		& \text{if } \zeta = \Omega(\beta^t_{a}(i,j), \beta^t_{hv}(i,j), \beta^t_{d}(i,j))\\
0,      & \text{otherwise}
\end{cases}
\end{equation}
where $\Omega$ is used to decode the three binary input values into a decimal value as follows for three binary inputs $b_1$, $b_2$, and $b_3$,
\begin{equation}
\Omega(b_1, b_2, b_3) = (b_1 \times 4) + (b_2 \times 2) + b_3 + 1.
\label{eq:Omega}
\end{equation}
Similarly, the $\Psi^{\zeta, t}_2 (i,j)$ for $2^{nd}$ decoder is calculated as follows,
\begin{equation}
\Psi^{\zeta, t}_2 (i,j) = 
\begin{cases}
1,		& \text{if } \zeta = \Omega(\beta^t_{a}(i,j), \beta^t_{sv}(i,j), \beta^t_{sh}(i,j))\\
0,      & \text{otherwise}
\end{cases}
\end{equation}
where $\Omega$ is defined in Eq. (\ref{eq:Omega}).
After finding the binary values of output channels for both decoders, i.e., $\Psi^{\zeta, t}_1 (i,j)$ and $\Psi^{\zeta, t}_2 (i,j)$ for $\zeta \in [1, 2^\gamma]$, $\gamma=3$ and $t \in [1, N]$, the output values, i.e., $\Psi^{\zeta}_1 (i,j)$ and $\Psi^{\zeta}_2 (i,j)$ are computed using Eq. (\ref{eq:output}).

The concept of frequency decoder is illustrated in Fig. \ref{decoder-example} with the help of an example face image taken from the LFW database \cite{lfw}. The Fig. \ref{decoder-example}a shows the example face image. The output filtered images are displayed in Fig. \ref{decoder-example}b by applying the $F_{hv}$, $F_{d}$, $F_{a}$, $F_{sv}$, and $F_{sh}$ filters over the input example face. Next, the LBP binary codes are generated over each filtered image in the Fig. \ref{decoder-example}c by using the algorithm presented in sub-section "LBP Binary Code Generation". Two decoders with three inputs each are used in Fig. \ref{decoder-example}d. The three inputs for the first decoder are LBP binary codes obtained over the filtered images generated using $F_{hv}$, $F_{d}$, and $F_{a}$ filters, whereas the inputs for the second decoder are LBP binary codes obtained over the filtered images generated using $F_{a}$, $F_{sv}$, and $F_{sh}$ filters. Finally, Fig. \ref{decoder-example}e presents the output images of the frequency decoders.

\begin{figure*}[!t]
    \centering
    \includegraphics[width=.99\linewidth]{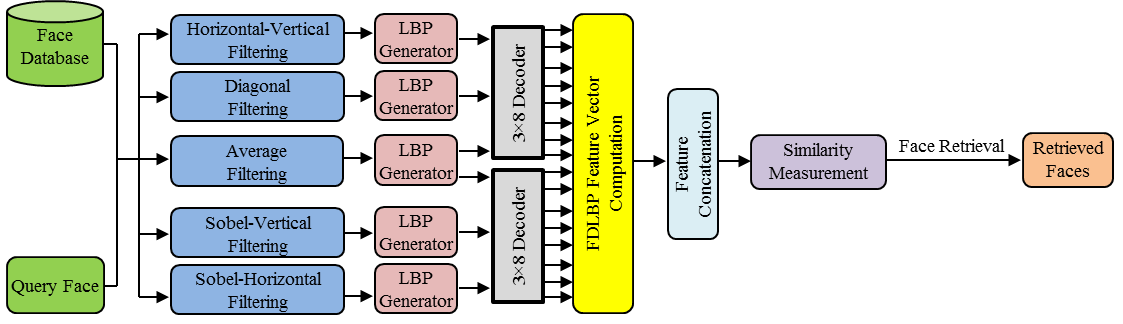}
    \caption{The face retrieval framework using the proposed FDLBP descriptor.}
    \label{fig:setup}
\end{figure*}

\subsection{Feature Vector Computation}
After finding the face representations by applying the frequency decoder over the multiple filtered images, the next task is to convert the representations into a feature vector form. The standard histogram strategy is used to achieve it. Suppose the final feature vector is denoted by $\xi$ and given as follows,
\begin{equation}
\begin{aligned}
\xi & = <\xi_1, \xi_2> \\
	& = <_{\eta = 1}^{2} \xi_\eta>
\end{aligned}
\end{equation}
where the $<\upsilon_1, \upsilon_2>$ symbol is used for the concatenation of the two vectors $\upsilon_1$ and $\upsilon_2$, and $\xi_1$ and $\xi_2$ are the feature vectors computed by using the output images of $1^{st}$ and $2^{nd}$ decoder respectively. The feature vector for the $\eta^{th}$ decoder, i.e., $\xi_\eta$ is computed as follows,
\begin{equation}
\begin{aligned}
\xi_\eta & = <\xi_\eta^{1}, \xi_\eta^{2}, ...., \xi_\eta^{\zeta}, ...., \xi_\eta^{2^\gamma}> \\
& = <_{\zeta = 1}^{2^\gamma}\xi_\eta^{\zeta}>
\end{aligned}
\end{equation}
where $\xi_\eta^{\zeta}$ is the feature vector generated as the histogram over the $\zeta^{th}$ output image of $\eta^{th}$ decoder. Mathematically, the $\xi_\eta^{\zeta}$ is calculated as follows,
\begin{equation}
\begin{aligned}
\xi_\eta^{\zeta} & = <\xi_\eta^{\zeta}(1), \xi_\eta^{\zeta}(2), ..., \xi_\eta^{\zeta}(\lambda), ...., \xi_\eta^{\zeta}(2^N-1)> \\
& = <_{\lambda=1}^{2^N-1}\xi_\eta^{\zeta}(\lambda)>
\end{aligned}
\end{equation}
where $N$ is the number of neighbors initially considered  to get the LBP binary codes, $\lambda$ is the bin number, and $\xi_\eta^{\zeta}(\lambda)$ is the number of occurrences of $\lambda$ in the output image $\xi_\eta^{\zeta}$ as follows,
\begin{equation}
\xi_\eta^{\zeta}(\lambda) = \sum_{i=R+1}^{x-R} \sum_{j=R+1}^{y-R} \rho(\lambda, \Psi_\eta^\zeta(i,j))
\end{equation}
where $R$ is the radius of local neighborhood considered initially for LBP binary code generation, $x \times y$ is the dimension of the image, and $\rho$ is a comparator function to check that two values are same or not. The $\rho$ for inputs $\alpha_1$ and $\alpha_2$ is defined by the following equation,
\begin{equation}
\rho(\alpha_1, \alpha_2) = 
\begin{cases}
1,		& \text{if } \alpha_1 = \alpha_2,\\
0,      & \text{otherwise}.
\end{cases}
\end{equation}

Suppose, the dimension of the feature vector of the $\zeta^{th}$ output image of $\eta^{th}$ decoder is represented by $D_{\eta}^{\zeta}$ and given as follows,
\begin{equation}
D_{\eta}^{\zeta} = {(2)}^{(N)}-1
\end{equation}
where $\zeta \in [1,2^\gamma]$, $\eta \in [1, \Gamma]$, $\gamma$ is the number of inputs to the decoder, $\Gamma$ is the number of decoders, and $N$ is the number of local neighbors of LBP binary code. The dimension of the feature vector computed using $\eta^{th}$ decoder, i.e., $D_{\eta}$ is computed as follows,
\begin{equation}
D_{\eta} = (2)^{(\gamma)} \times ({(2)}^{(N)}-1)
\end{equation}
Similarly, the dimension of final feature vector ($D$) is given as,
\begin{equation}
D = \Gamma \times (2)^{(\gamma)} \times ({(2)}^{(N)}-1)
\end{equation}
In this work, the default values of $\Gamma$, $\gamma$ and $N$ are $2$, $3$ and $8$ respectively. Thus, the dimension of the final feature vector is $2 \times (2)^{(3)} \times ((2)^{(8)}-1) = 2 \times 8 \times 256 = 16 \times 256$.

In order to make the feature vector invariant against the image resolution, it is normalized as follows,
\begin{equation}
\xi_{norm}(\mu) = \frac{\xi(\mu)}{\sum_{\tau = 1}^{D} \xi(\tau)}
\end{equation}
for $\forall$ $\mu \in [1, D]$. The normalized version of the feature vectors is used in the experiments.

\subsection{Color FDLBP Descriptor}
In the previous sub-sections the frequency decoded local binary pattern (FDLBP) feature vector is computed over the grayscale image. The mdLBP is proposed for the color images \cite{mdlbp}. In order to show the significance of the frequency decoder over color channel decoder, the color FDLBP (i.e., cFDLBP) and the frequency mdLBP (i.e., FmdLBP) are also proposed in this paper. The cFDLBP computes the FDLBP feature vector over each color channel independently. Thus, six frequency decoder is used with two per color channel. Finally, FDLBP feature vectors of all frequency decoders are concatenated to find the single cFDLBP feature vector. Whereas, on the other hand, the FmdLBP uses the color decoder of original mdLBP \cite{mdlbp}. A total of five color decoders corresponding to the five filters are used with three inputs each. The three inputs to a color decoder are from the three color channels respectively filtered with a particular filter.

\section{Experimental Setup}
The experiments using the framework of the image retrieval are conducted in this paper. The face retrieval using the the proposed frequency decoded local binary pattern (FDLBP) descriptor is depicted in the Fig. \ref{fig:setup}. For a query face image, the top matching face images are retrieved from a database using the FDLBP descriptor. The FDLBP descriptor is computed over the query face image and the all database face images. The similarity scores are calculated between the LDRP descriptors of the query and the database images with the help of the some distance measure techniques. Note that the high similarity score (i.e., low distance) between two descriptors indicates that the corresponding faces are more similar and vice-versa. 

\subsection{Distances Measures}
For image matching, the performance is highly dependent upon the distance measures used for finding the similarity scores. Based on the best similarity scores, the top $n$ number of faces are retrieved. In the experiments, the Euclidean, Cosine, Earth Mover Distance (Emd), L1, D1, and Chi-square distances are used \cite{ltrp}, \cite{mdlbp}. The Chi-square distance is used as the default similarity measure in this paper.

\subsection{Evaluation Criteria}
The precision, recall, f-score, and retrieval rank metrics are generally used to evaluate the image retrieval algorithms. In order to judge the performance over a database, the metrics are computed by converting all the images of that database as the query image one by one. The average retrieval precision (ARP) and average retrieval rate (ARR) over whole database are calculated by taking the average of mean precision (MP) and mean recall (MR) of each category of that database respectively. The MP and MR for a category are computed by taking the mean of precision and recall by turning all of the images in that category as the query image one by one respectively. The precision ($Pr$) and recall ($Re$) for a query image is calculated as follows,
\begin{equation}
\begin{aligned}
& Pr = \frac{\#Correct\_Retrieved\_Images}{\#Retrieved\_Images}\\
& Re = \frac{\#Correct\_Retrieved\_Images}{\#Similar\_Images\_In\_Database}
\end{aligned}
\end{equation}
The F-score is computed from the ARP and ARR values as follows,
\begin{equation}
F-score=2 \times \frac{ARP\times ARR}{ARP+ARR}
\end{equation}
The average normalized modified retrieval rank (ANMRR) metric is also used to test the rank of correctly retrieved faces \cite{anmrr}. The higher value of ARP, ARR and F-Score indicates the better retrieval performance and vice-versa, whereas the lower value of ANMRR indicates the better retrieval performance and vice-versa.

\begin{figure*}[!t]  
  \begin{subfigure}{.5\textwidth}
    \centering
    \includegraphics[width=.98\linewidth]{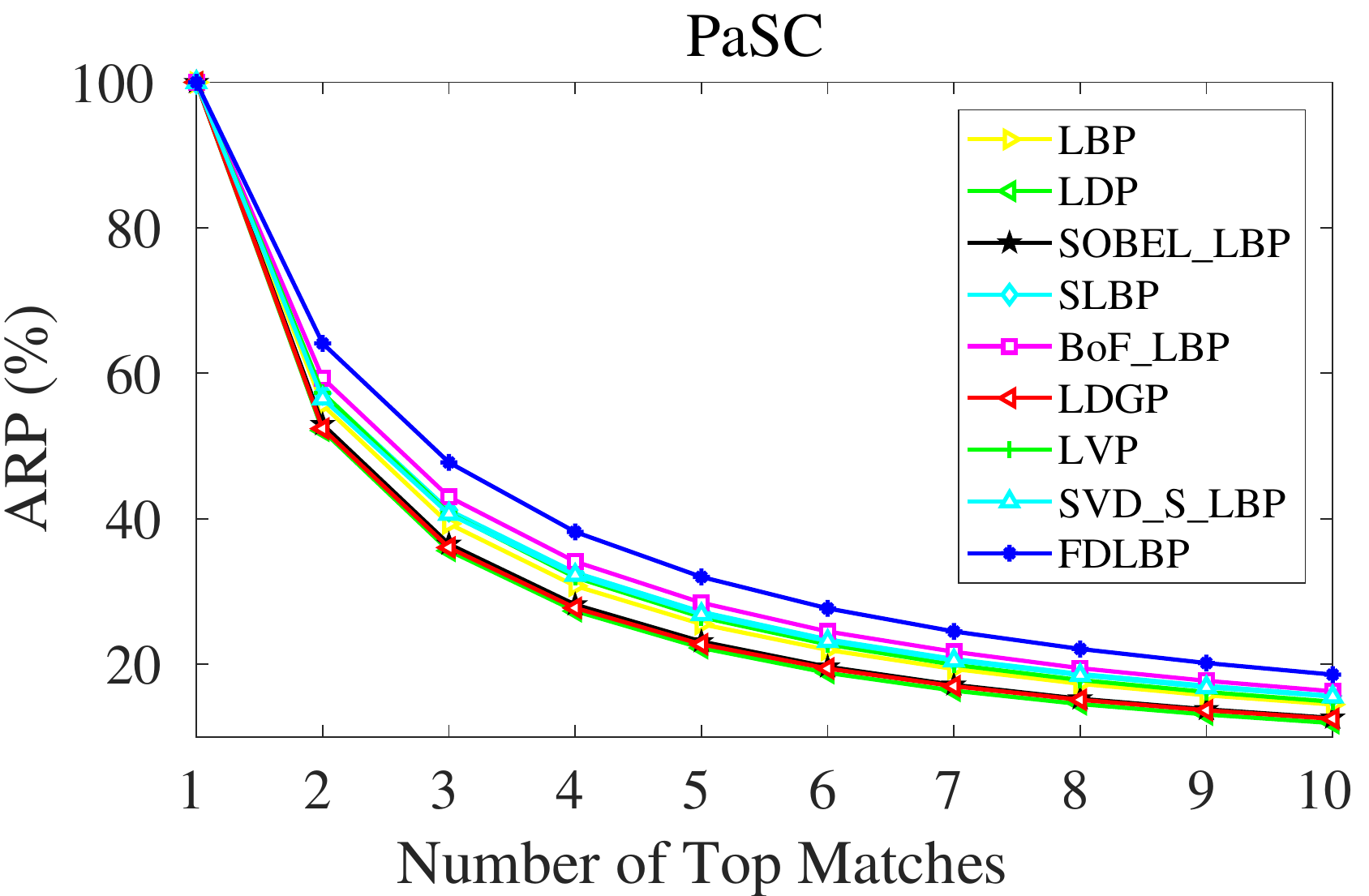}
    \caption{ARP}
    \label{fig:pasc-arp}
  \end{subfigure}%
    \begin{subfigure}{.5\textwidth}
    \centering
    \includegraphics[width=.98\linewidth]{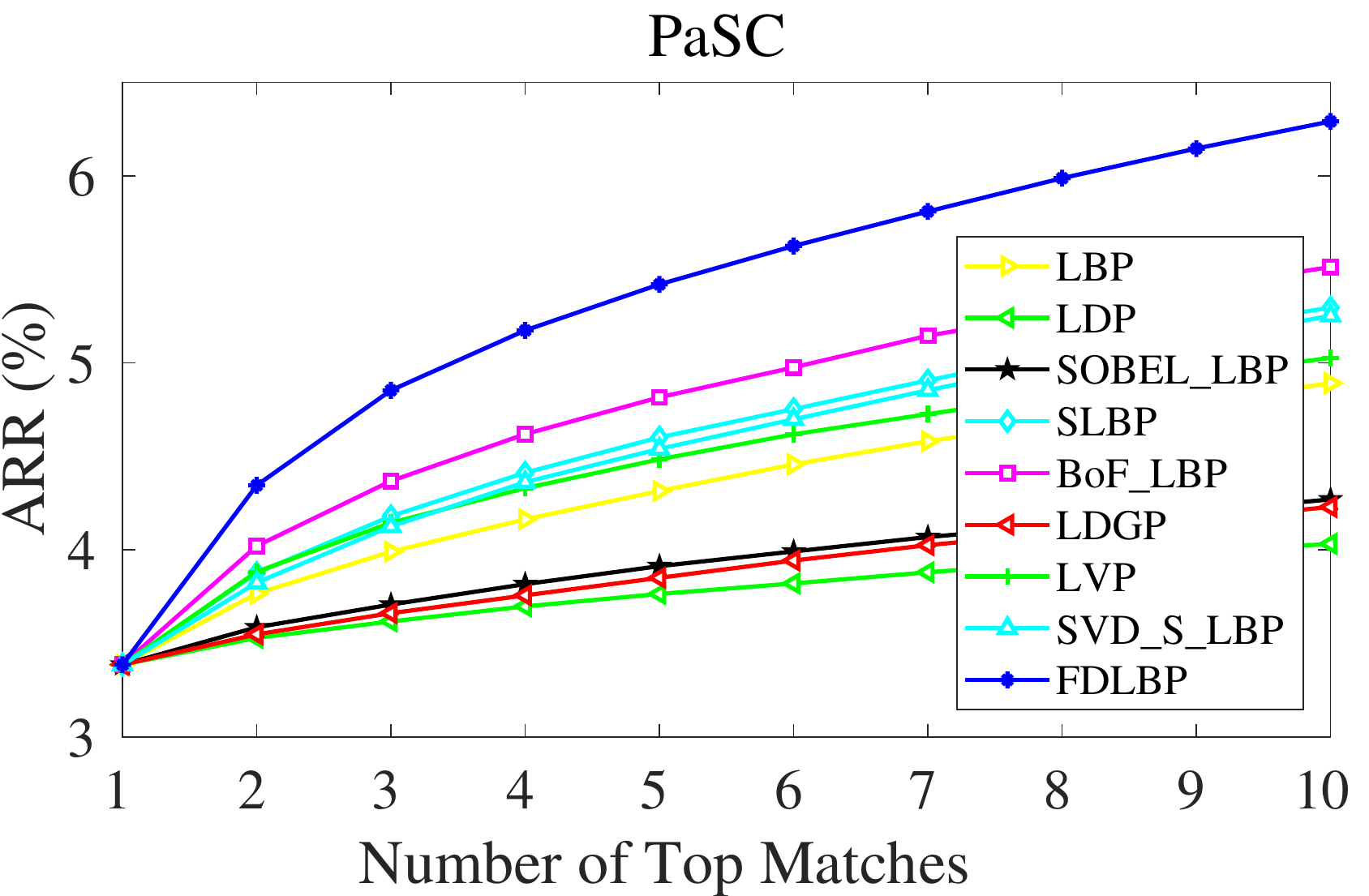}
    \caption{ARR}
    \label{fig:pasc-arr}
  \end{subfigure}
    \begin{subfigure}{.5\textwidth}
    \centering
    \includegraphics[width=.98\linewidth]{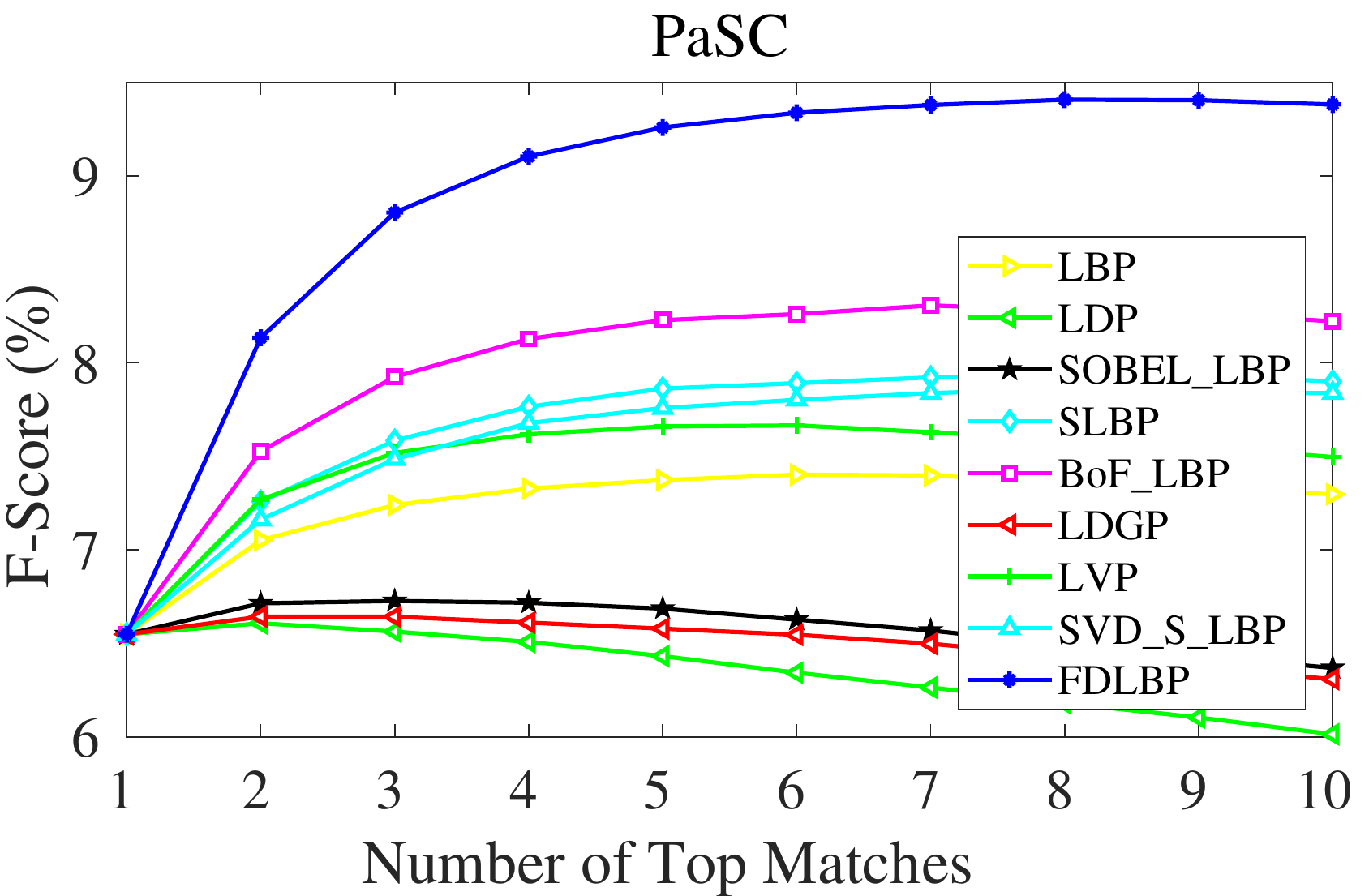}
    \caption{F-Score}
    \label{fig:pasc-f}
  \end{subfigure}%
    \begin{subfigure}{.5\textwidth}
    \centering
    \includegraphics[width=.98\linewidth]{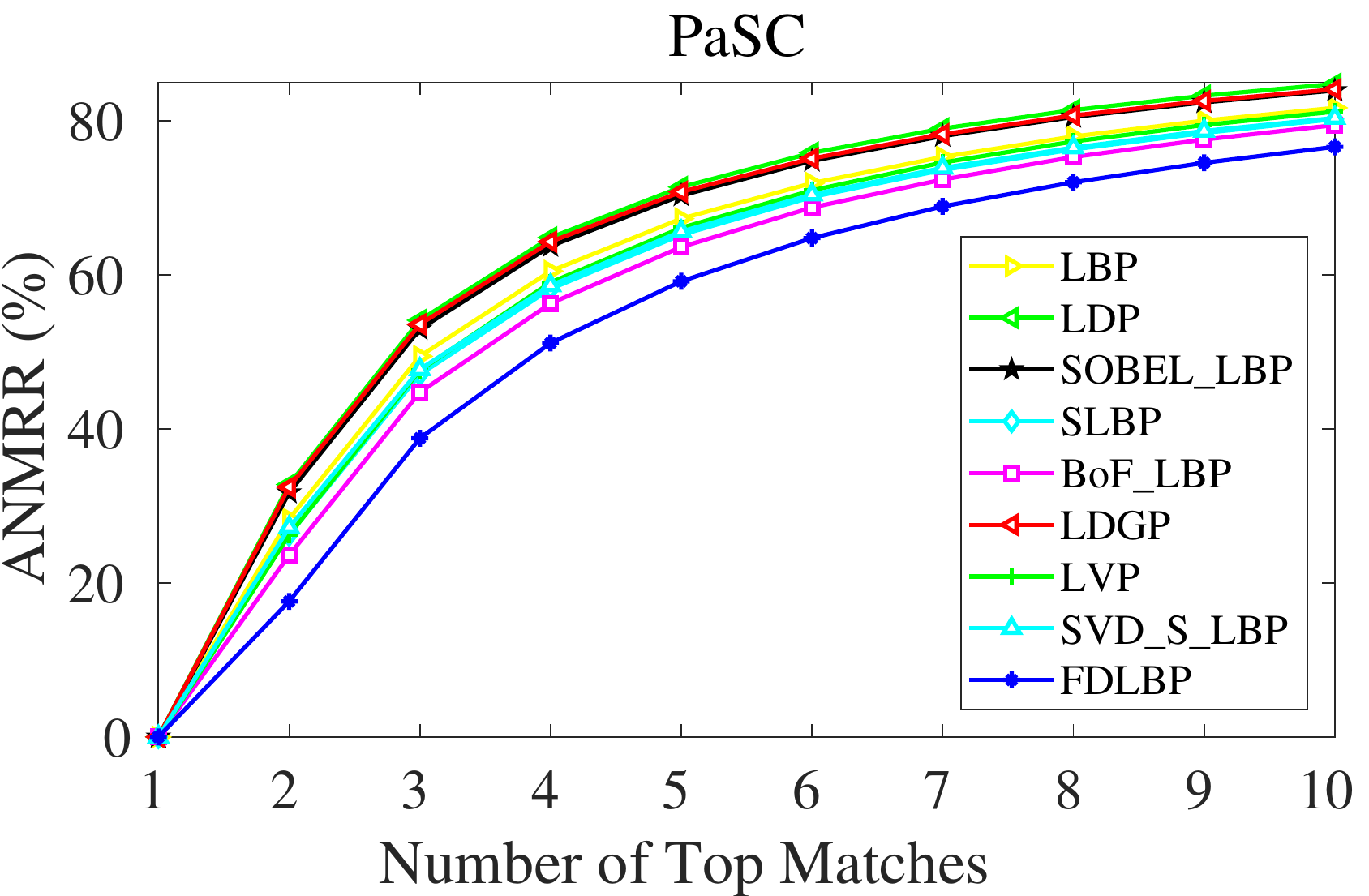}
    \caption{ANMRR}
    \label{fig:pasc-anmrr}
  \end{subfigure}
  \caption{The results over PaSC face database in terms of the (a) ARP, (b) ARR, (c) F-Score, and (d) ANMRR vs number of retrieved images. The x-axis shows the number of retrieved images ($n$). The y-axis represents the ARP, ARR, F-score, and ANMRR metric values.}
  \label{fig:results-pasc}
\end{figure*}

\begin{figure*}[!t]  
  \begin{subfigure}{.5\textwidth}
    \centering
    \includegraphics[width=.98\linewidth]{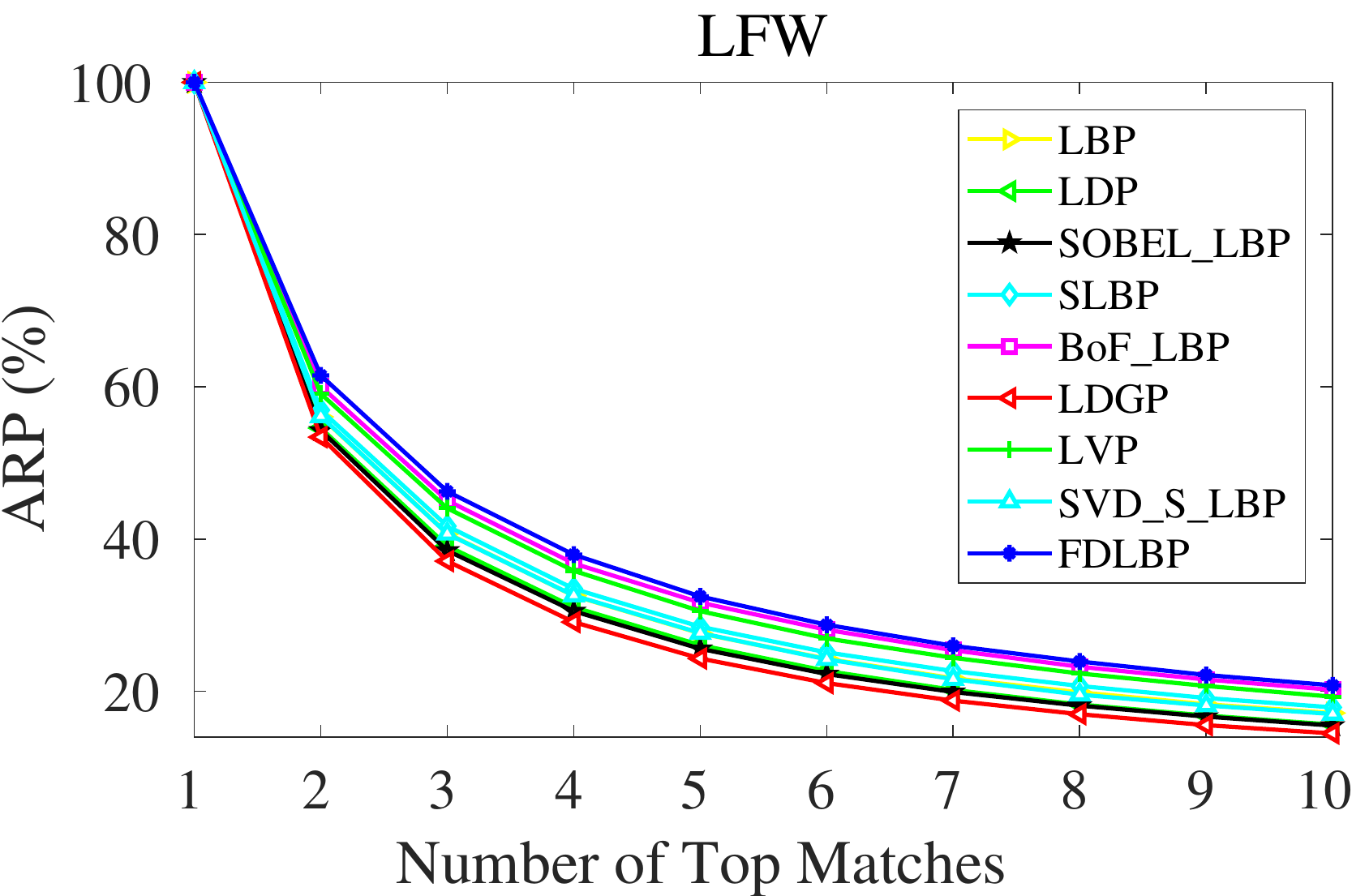}
    \caption{ARP}
    \label{fig:lfw-arp}
  \end{subfigure}%
    \begin{subfigure}{.5\textwidth}
    \centering
    \includegraphics[width=.98\linewidth]{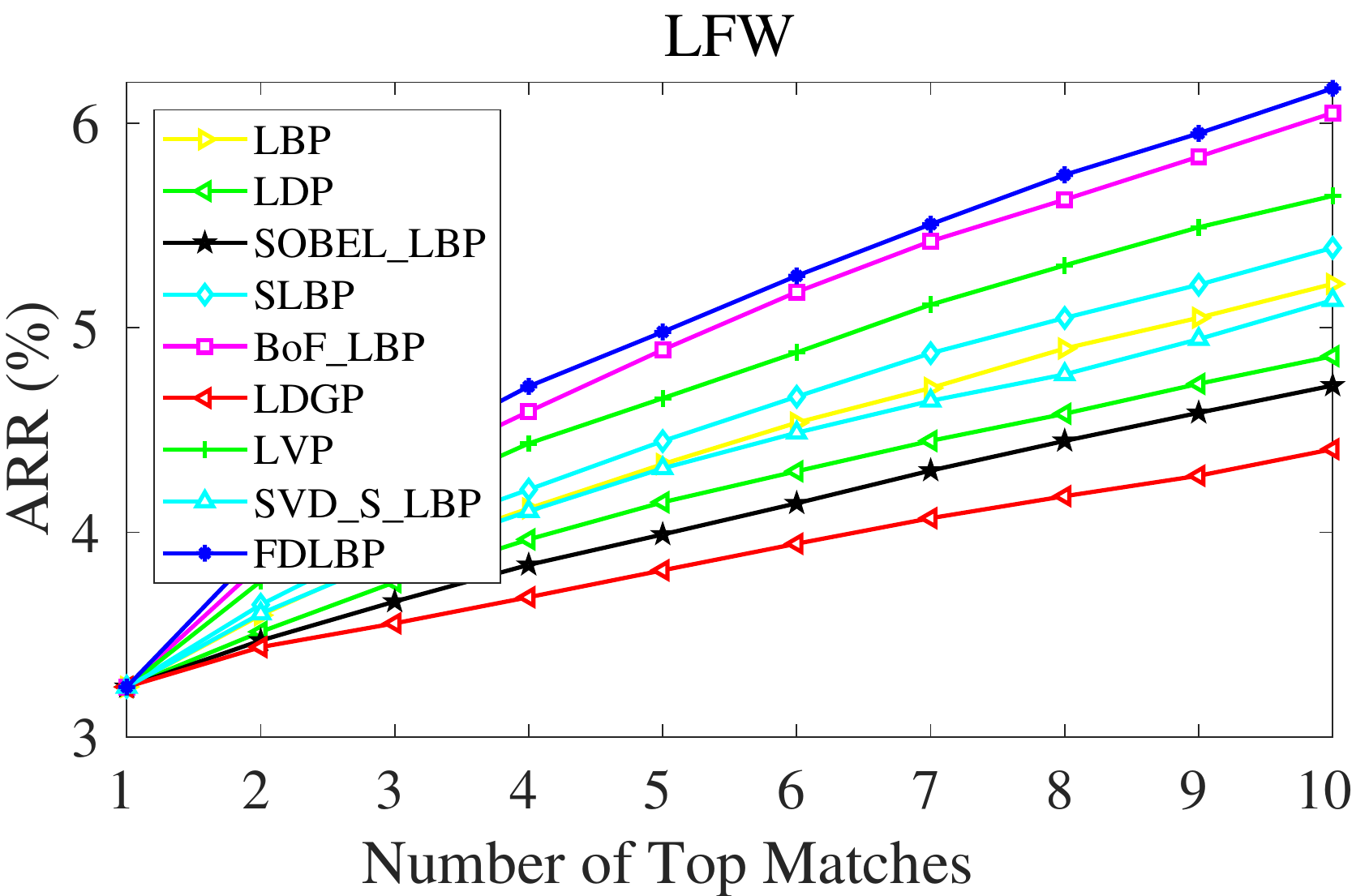}
    \caption{ARR}
    \label{fig:lfw-arr}
  \end{subfigure}
    \begin{subfigure}{.5\textwidth}
    \centering
    \includegraphics[width=.98\linewidth]{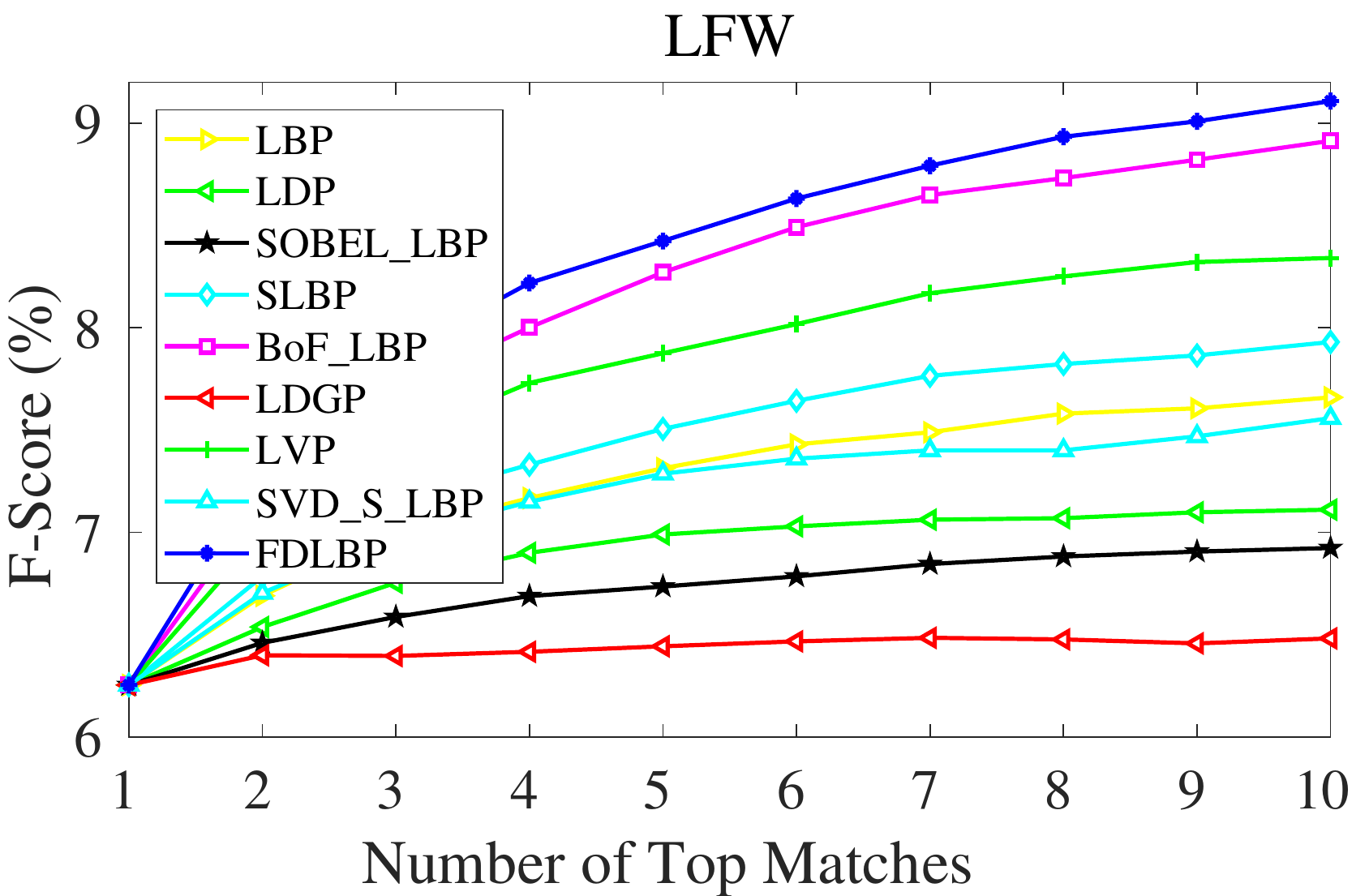}
    \caption{F-Score}
    \label{fig:lfw-f}
  \end{subfigure}%
    \begin{subfigure}{.5\textwidth}
    \centering
    \includegraphics[width=.98\linewidth]{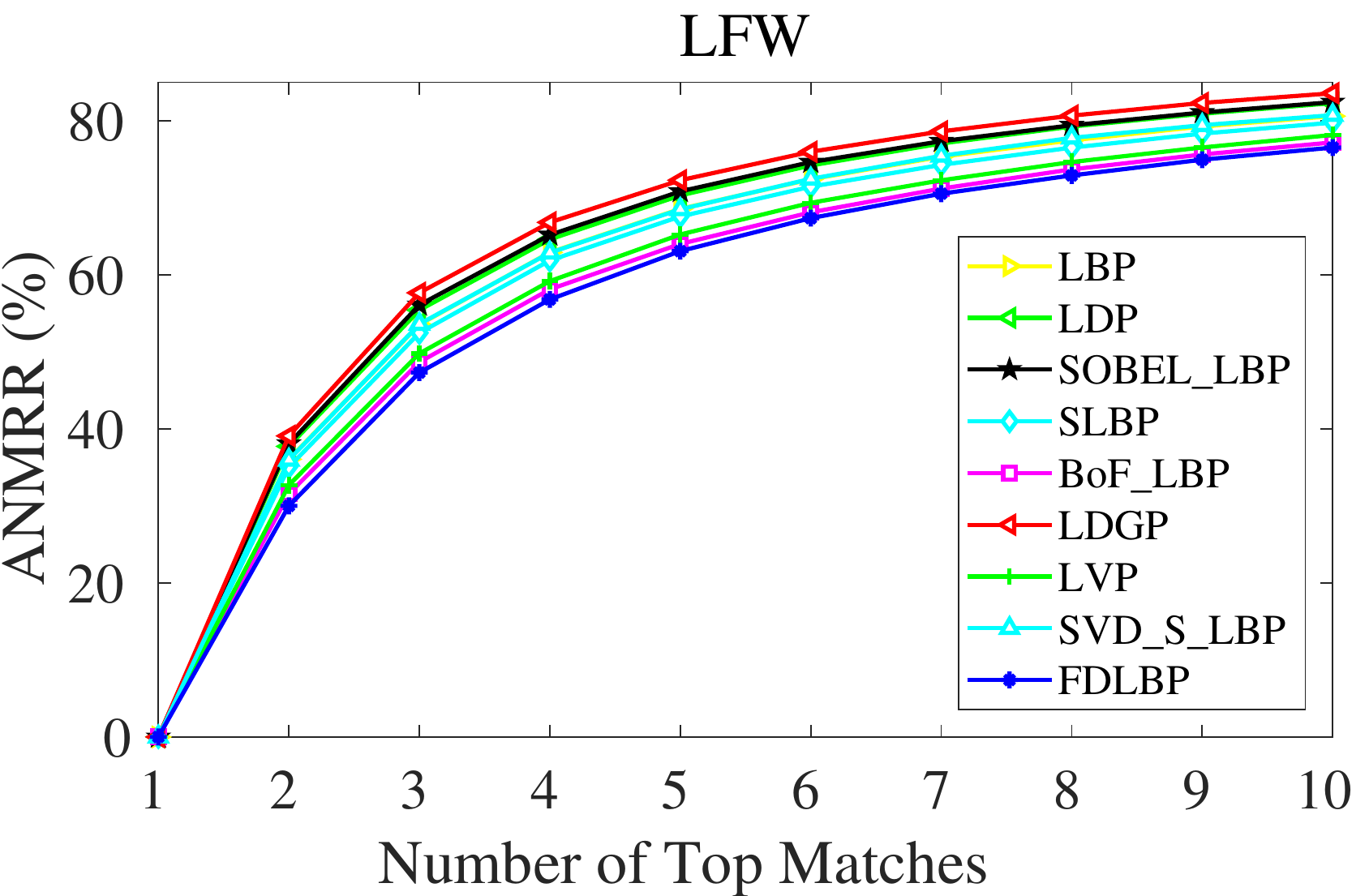}
    \caption{ANMRR}
    \label{fig:lfw-anmrr}
  \end{subfigure}
  \caption{The results over LFW face database in terms of the (a) ARP, (b) ARR, (c) F-Score, and (d) ANMRR vs number of retrieved images. The x-axis shows the number of retrieved images ($n$). The y-axis represents the ARP, ARR, F-score, and ANMRR metric values.}
  \label{fig:results-lfw}
\end{figure*}

\begin{figure*}[!t]  
  \begin{subfigure}{.5\textwidth}
    \centering
    \includegraphics[width=.98\linewidth]{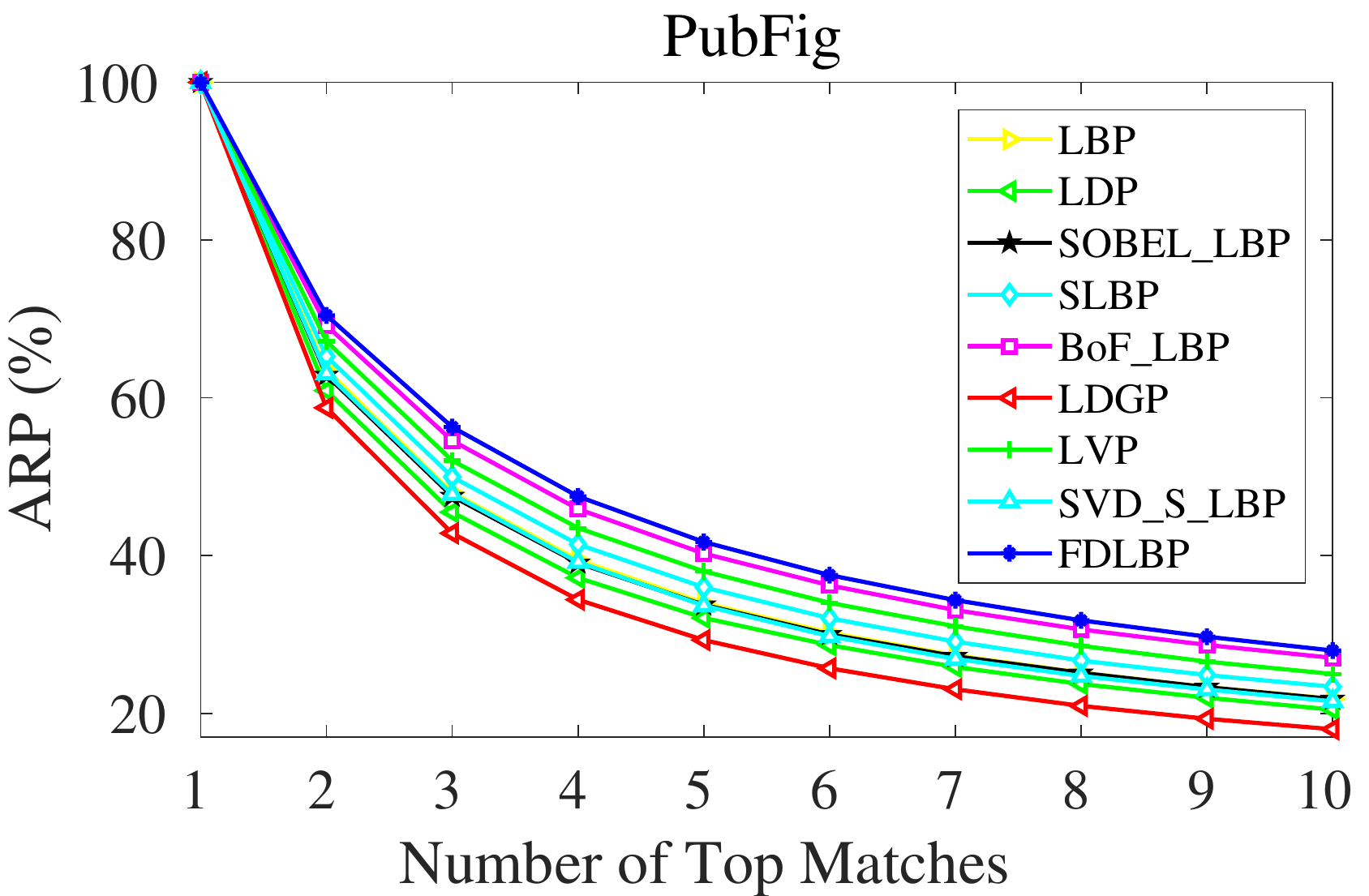}
    \caption{ARP}
    \label{fig:pubfig-arp}
  \end{subfigure}%
    \begin{subfigure}{.5\textwidth}
    \centering
    \includegraphics[width=.98\linewidth]{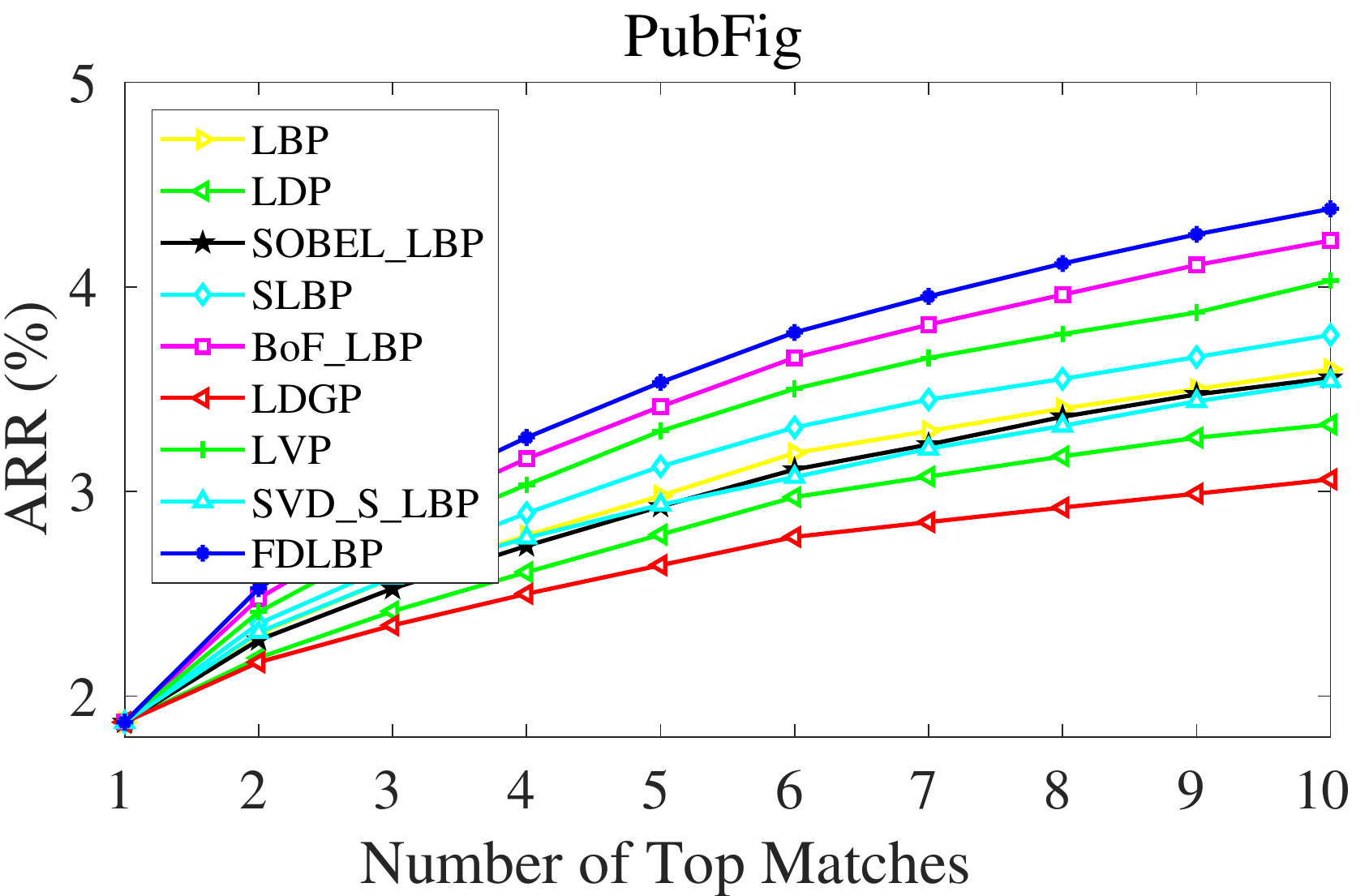}
    \caption{ARR}
    \label{fig:pubfig-arr}
  \end{subfigure}
    \begin{subfigure}{.5\textwidth}
    \centering
    \includegraphics[width=.98\linewidth]{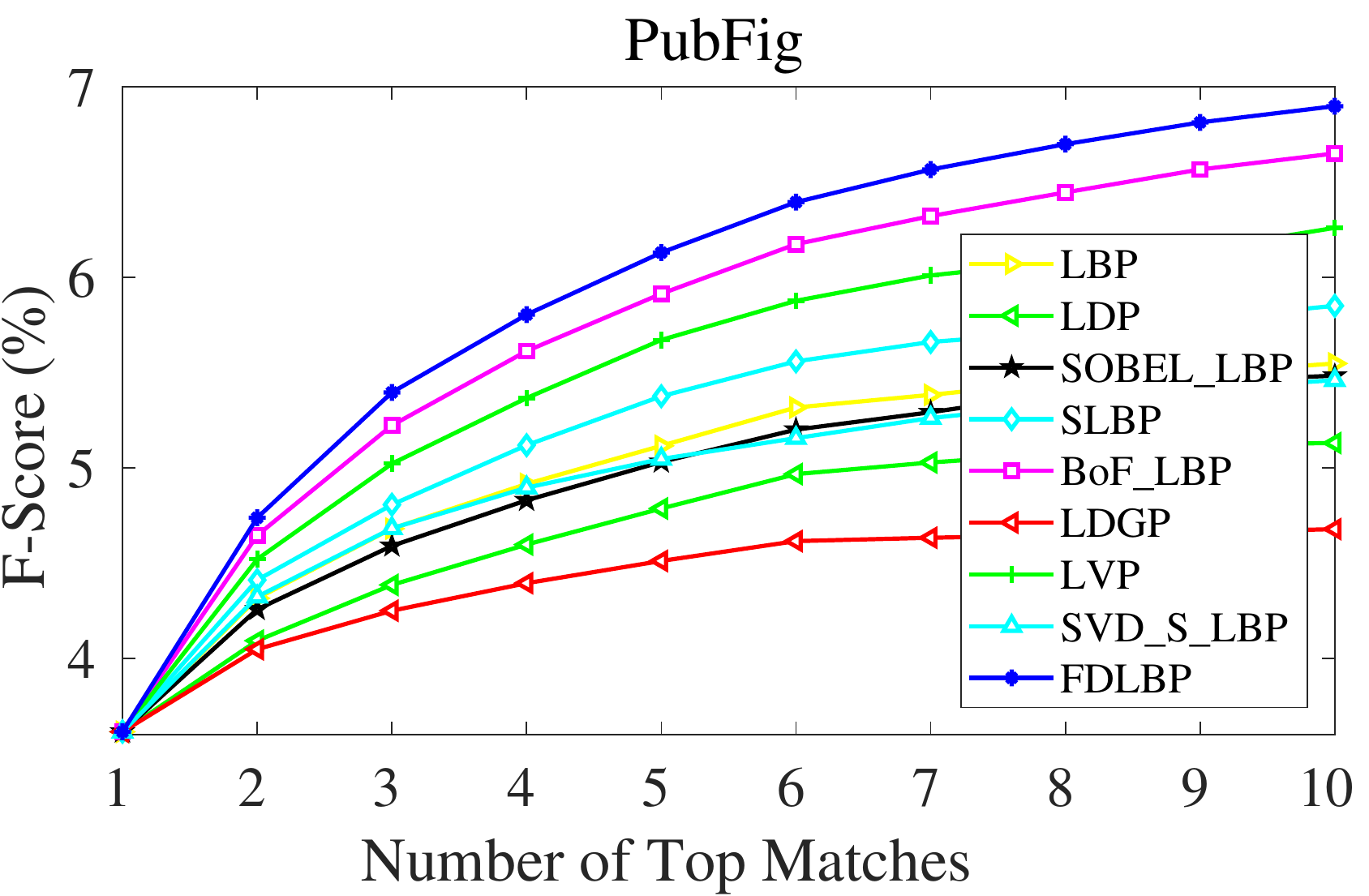}
    \caption{F-Score}
    \label{fig:pubfig-f}
  \end{subfigure}%
    \begin{subfigure}{.5\textwidth}
    \centering
    \includegraphics[width=.98\linewidth]{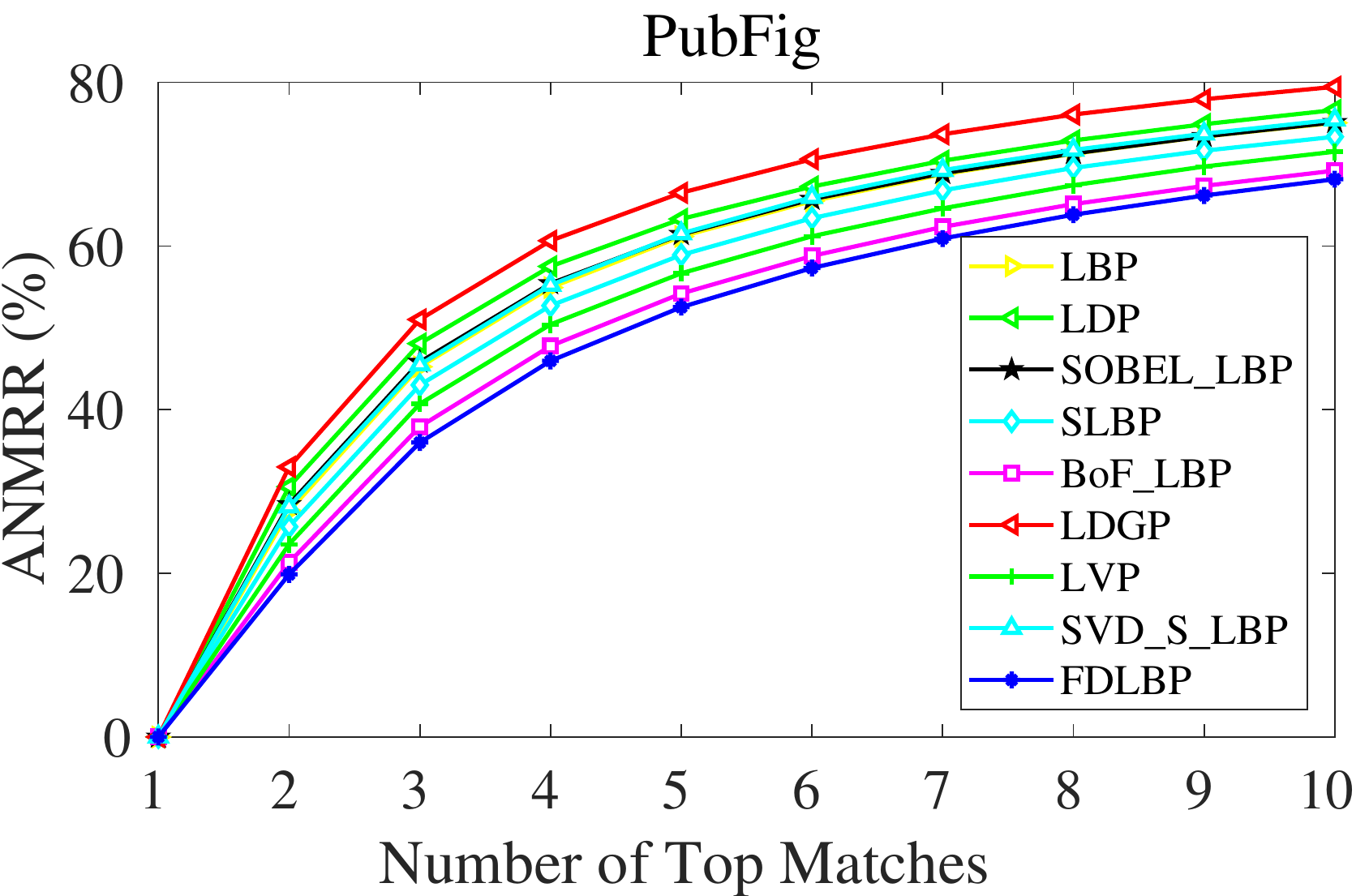}
    \caption{ANMRR}
    \label{fig:pubfig-anmrr}
  \end{subfigure}
  \caption{The results over PubFig face database in terms of the (a) ARP, (b) ARR, (c) F-Score, and (d) ANMRR vs number of retrieved images. The x-axis shows the number of retrieved images ($n$). The y-axis represents the ARP, ARR, F-score, and ANMRR metric values.}
  \label{fig:results-pubfig}
\end{figure*}

\begin{figure*}[!t]  
 \begin{subfigure}{.5\textwidth}
    \centering
    \includegraphics[width=.98\linewidth]{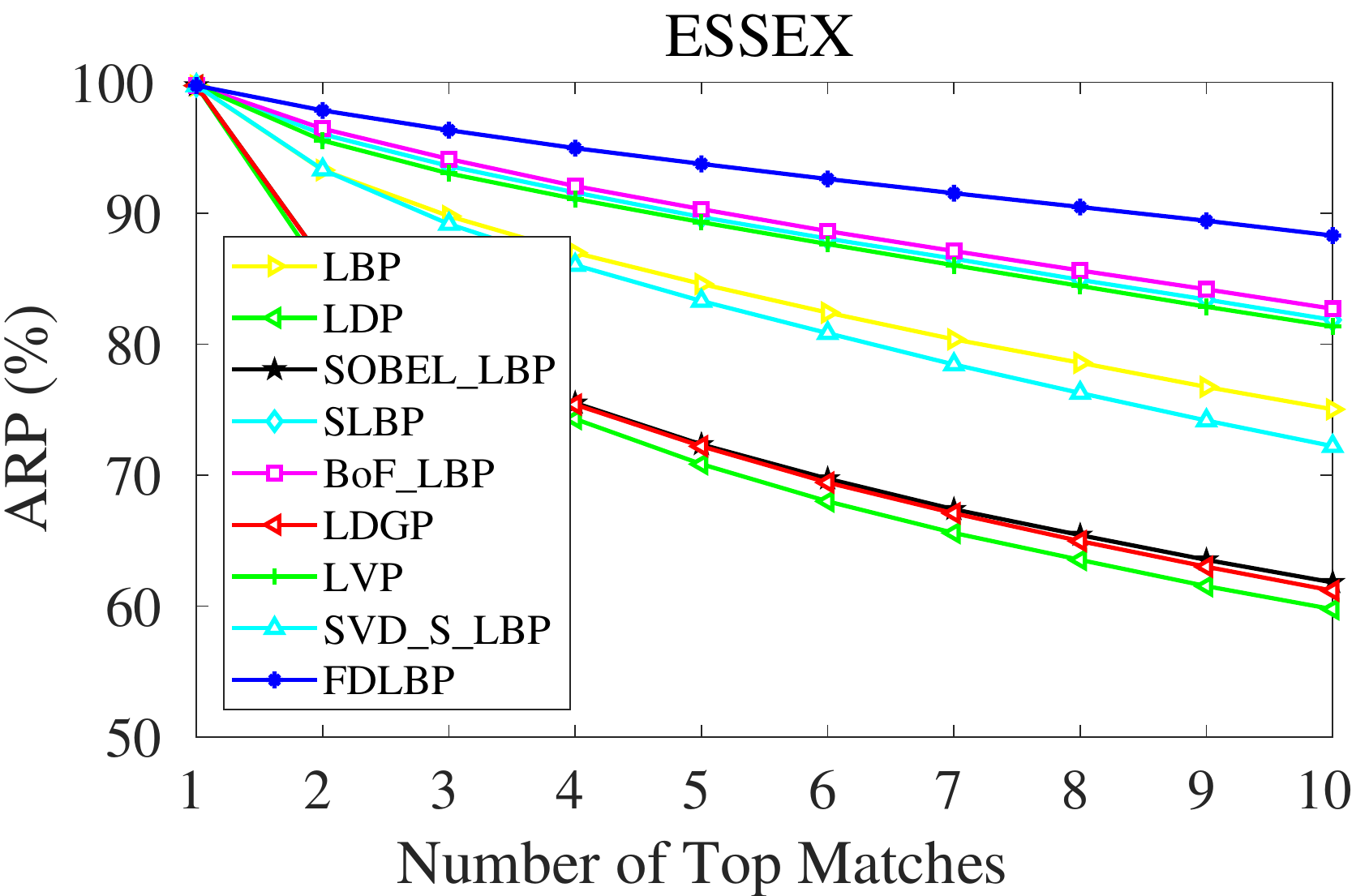}
    \caption{ARP}
    \label{fig:essex-arp}
  \end{subfigure}%
    \begin{subfigure}{.5\textwidth}
    \centering
    \includegraphics[width=.98\linewidth]{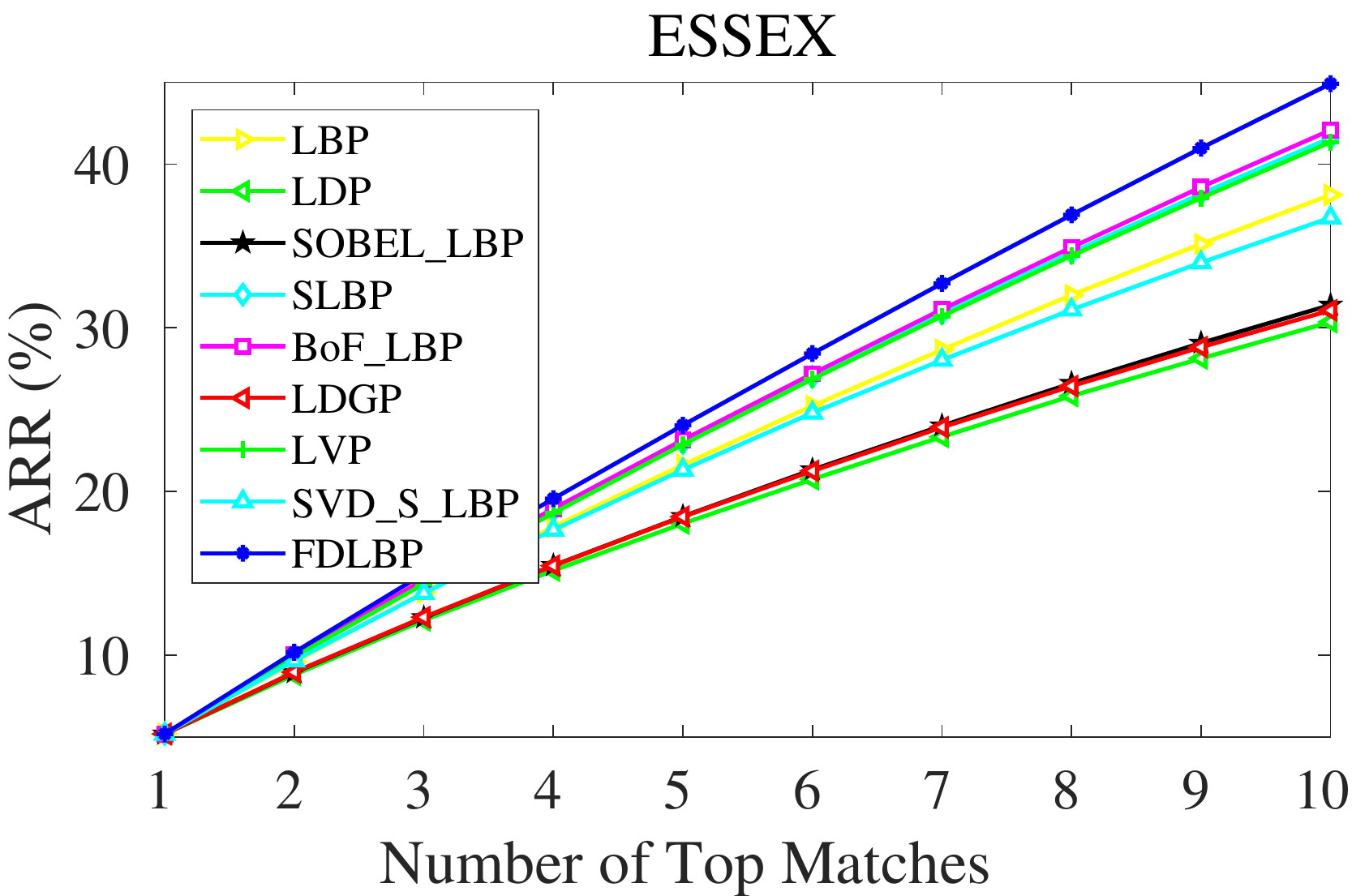}
    \caption{ARR}
    \label{fig:essex-arr}
  \end{subfigure}
    \begin{subfigure}{.5\textwidth}
    \centering
    \includegraphics[width=.98\linewidth]{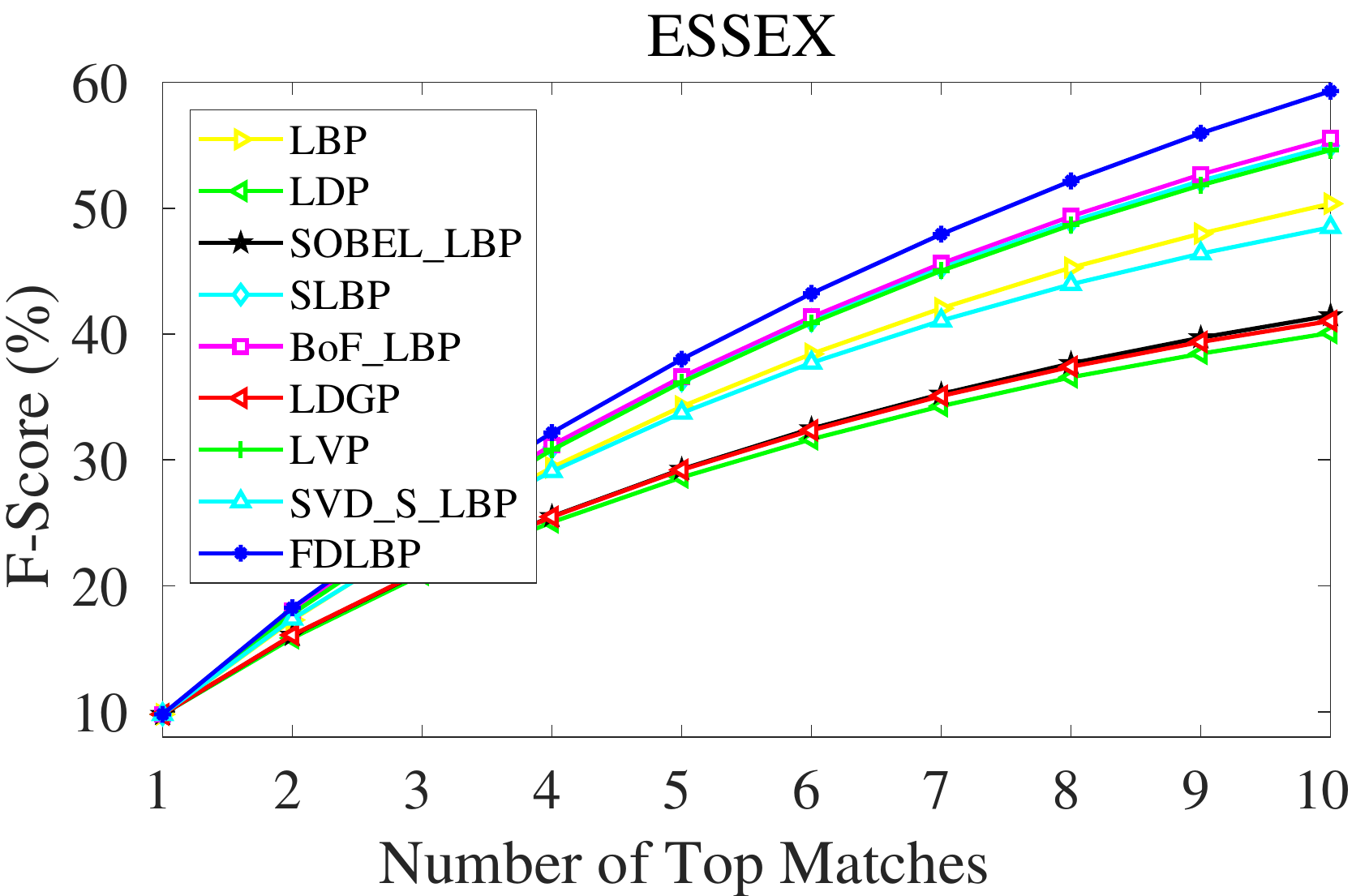}
    \caption{F-Score}
    \label{fig:essex-f}
  \end{subfigure}%
    \begin{subfigure}{.5\textwidth}
    \centering
    \includegraphics[width=.98\linewidth]{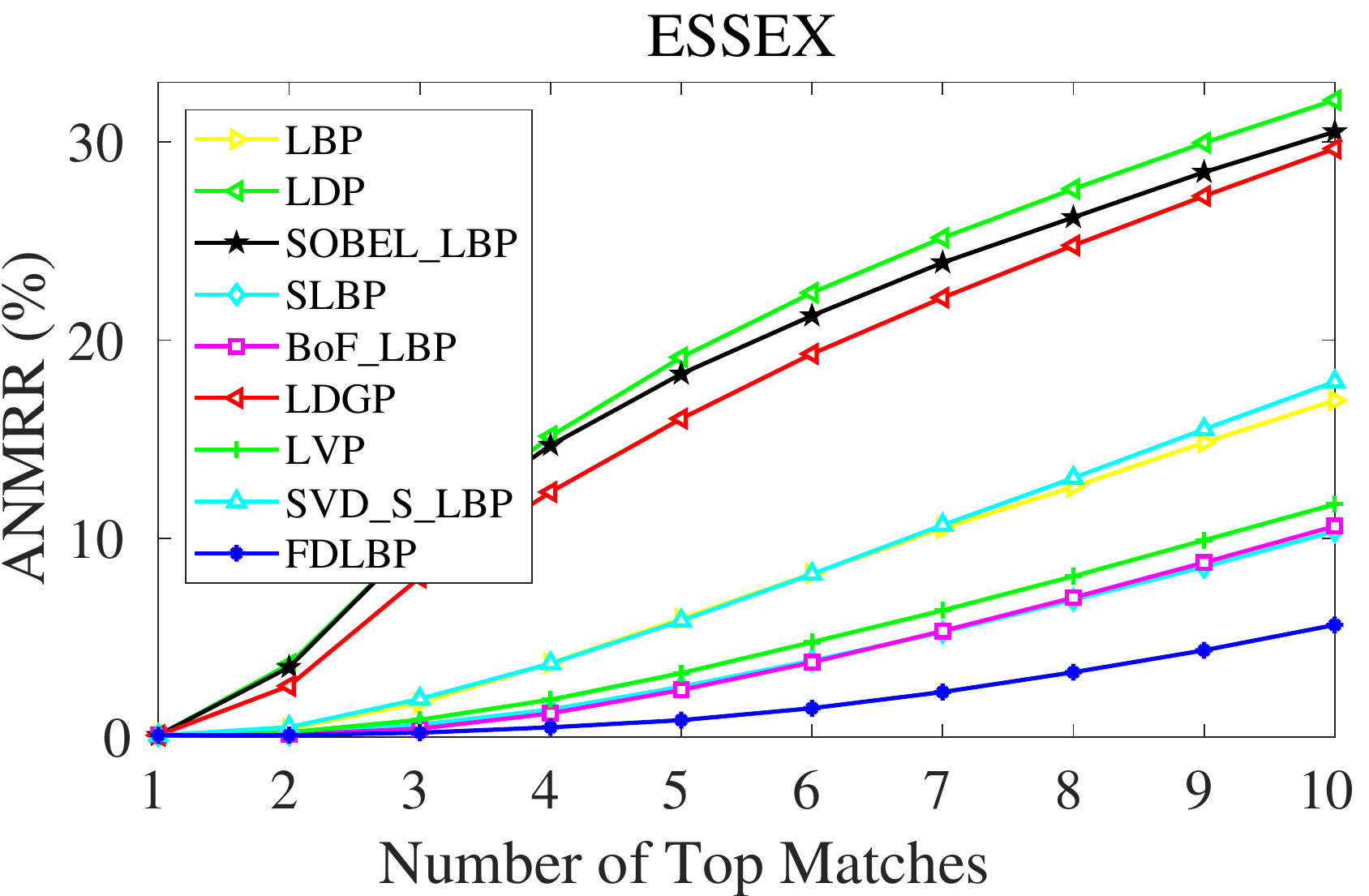}
    \caption{ANMRR}
    \label{fig:essex-anmrr}
  \end{subfigure}
  \caption{The results over ESSEX face database in terms of the (a) ARP, (b) ARR, (c) F-Score, and (d) ANMRR vs number of retrieved images. The x-axis shows the number of retrieved images ($n$). The y-axis represents the ARP, ARR, F-score, and ANMRR metric values.}
  \label{fig:results-essex}
\end{figure*}

\subsection{Databases Used}
In order to judge the improved performance of the proposed descriptor, we have used four challenging face databases namely PaSC \cite{pasc}, LFW \cite{lfw}, PubFig \cite{pubfig}, and ESSEX \cite{essex}. The color images are converted into the grayscale except for the color descriptors. The Viola Jones object detection method is used to localize the face regions in the image  \cite{viola}. The cropped face images are down-sampled to $64\times64$ resolution. 
\subsubsection{PaSC Face Database \cite{pasc}}
The PaSC still images face database is one of the appealing databases and has the challenges like blur, pose, and illumination \cite{pasc}. A total of 8718 faces are detected using Viola Jones detector belonging to 293 subjects in the PaSC database. 
\subsubsection{LFW Face Database \cite{lfw}}
The unconstrained face retrieval is very challenging and desirable in the current scenario. The LFW database consists the images from the Internet \cite{lfw}. These images are captured without the cooperations from subjects. The variations like pose, lighting, expression, scene, camera, etc. are present in this database. In the image retrieval framework, it is required to retrieve more than one (typically 5, 10, etc.) best matching images. So, it is important that the sufficient number of images must be available in each category of the database. Thus, the subjects having at least 20 images in LFW database are considered. A total of 2984 face images from 62 individuals are present in the LFW database used in this paper. 
\subsubsection{PubFig Face Database \cite{pubfig}}
The Public Figure database (i.e., PubFig) is also an unconstrained database \cite{pubfig}. It contains the images from 60 subjects with 6472 number of total images \cite{pubfig}. The images are downloaded from the Internet directly following the non-dead urls given in this database. 
\subsubsection{ESSEX Face Database \cite{essex}}
The ESSEX face database is a very challenging database \cite{essex}. It consists many variations like background, scale, illumination, blur, and extreme variation of expressions \cite{essex}. A total of 7740 face images are present from 392 subjects with nearly 20 images per subject in ESSEX database.

\section{Face Retrieval Experimental Results}
In this section, the experimental results are reported and discussed. First of all, the retrieval results using the proposed descriptor are compared with the state-of-the-art descriptors. Second, the performance of the proposed frequency decoder is tested with the different combinations of the frequency decoder. Third, the suitability of the distance measures are experimented with the proposed descriptor. Forth, the color channels are used with the frequency decoder to test the performance over the color face images.

\subsection{Results Comparison}
The retrieval results are computed using ARP(\%), ARR(\%), F-score(\%) and ANMRR(\%) metrics over the four challenging constrained and unconstrained face databases including the PaSC, LFW, PubFig and ESSEX. The results of the proposed FDLBP descriptor are compared with the existing state-of-the-art descriptors such as the LBP \cite{lbp}, LDP \cite{ldp}, SOBEL\_LBP \cite{sobel-lbp}, SLBP \cite{slbp}, BoF\_LBP \cite{boflbp}, LDGP \cite{ldgp}, LVP \cite{lvp}, and SVD\_S\_LBP \cite{svdslbp}. Note that except the BoF\_LBP descriptor, rest of the descriptors are proposed for the face images. It is also worth to note that the SOBEL\_LBP, SLBP, BoF\_LBP and SVD\_S\_LBP descriptors are using some filters for the pre-processing.

The Fig. \ref{fig:results-pasc}, Fig. \ref{fig:results-lfw}, Fig. \ref{fig:results-pubfig}, and Fig. \ref{fig:results-essex} demonstrate the results over the PaSC, LFW, PubFig and ESSEX databases, respectively. The ARP ($1^{st}$ row, $1^{st}$ column), ARR ($1^{st}$ row, $2^{nd}$ column), F-score ($2^{nd}$ row, $1^{st}$ column) and ANMRR ($2^{nd}$ row, $2^{nd}$ column) metrics are plotted against the number of retrieved images ($n$). The x-axis shows the number of retrieved images ($n$), whereas the y-axis shows the ARP, ARR, F-score and ANMRR values in percentage. It is clear from the results that the FDLBP descriptor achieves the best result among all the descriptors as the values of ARP, ARR and F-score are highest and the value of ANMRR is lowest for the FDLBP descriptor. The PaSC and ESSEX are the constrained databases, but having many variations like blur, scale, illumination, expressions, and pose. The LFW and PubFig databases are having the unconstrained images. The results of Fig. \ref{fig:results-pasc}, Fig. \ref{fig:results-lfw}, Fig. \ref{fig:results-pubfig}, and Fig. \ref{fig:results-essex} confirm the improved performance of the proposed frequency decoder based FDLBP descriptor.

\begin{table*}[!t]
\caption{The ARP(\%) using the proposed FDLBP descriptor with different decoder combinations over the PaSC, LFW, PubFig, and ESSEX face databases. The number of retrieved images ($n$) is 5 in this experiment. The highest ARP values are expressed in bold for each database. The $<(F_a,F_{hv},F_{d})>$ represents that only one decoder is used in this case with three inputs as the filtered images obtained using $F_a$, $F_{hv}$, and $F_{d}$ filters respectively. The $<(F_{hv},F_{d}),(F_{sv},F_{sh})>$ represents that two decoders with two inputs each are used in this case with filtered images using $F_{hv}$ and $F_{d}$ filters in the first decoder and filtered images using $F_{hv}$ and $F_{d}$ filters in the second decoder.}  
\label{t1}
\begin{center}
\begin{tabular}{lcccc}
\hline
\\[-0.65em]\multirow{2}{*}{Decoder Combination} & \multicolumn{4}{c}{Face Databases}\\ \\[-0.65em]
\cline{2-5} 
\\[-0.65em] & PaSC & LFW & PubFig & ESSEX \\ \\[-0.65em]
\hline
\\[-0.65em]
$<(F_a,F_{hv},F_{d})>$ & 32.53 & 32.36 & 41.17 & \textbf{95.05}\\ \\[-0.85em]
$<(F_a,F_{sv},F_{sh})>$ & 28.92 & 30.02 & 38.22 & 90.32\\ \\[-0.85em]
$<(F_{hv},F_{d}),(F_{sv},F_{sh})>$ & 24.56 & 28.75 & 36.49 & 79.74\\ \\[-0.85em]
$<(F_{a},F_{hv}),(F_{a},F_{d}),(F_{a},F_{sv}),(F_{a},F_{sh})>$ & \textbf{32.59} & 32.31 & 41.52 & 94.15\\ \\[-0.85em]
$<(F_a,F_{hv},F_{sv}),(F_a,F_{d},F_{sh})>$ & 31.64 & \textbf{32.76} & 41.68 & 92.86\\ \\[-0.85em]
$<(F_a,F_{hv},F_{d}),(F_a,F_{sv},F_{sh})>$ & 32.00 & 32.46 & \textbf{41.73} & 93.77\\ \\[-0.65em]
\hline
\end{tabular}
\end{center}
\end{table*}

\subsection{Experiment with Different Decoder Combinations}
In the FDLBP descriptor, by default two decoders are used with three inputs each. The first decoder has the inputs as the outputs of $F_a$, $F_{hv}$ and $F_d$ filters and is represented in this subsection as $(F_a, F_{hv}, F_{d})$. Similarly the second decoder has the inputs coming from $F_a$, $F_{sv}$ and $F_{sh}$ filters and is represented as $(F_a, F_{sv}, F_{sh})$. So, in this subsection, $<(F_a, F_{hv}, F_{d}), (F_a, F_{sv}, F_{sh})>$ is the representation of the default decoder setting. For example, $<(F_{a},F_{hv}),(F_{a},F_{d}),(F_{a},F_{sv}),(F_{a},F_{sh})>$ represents four decoders with two inputs each. In this experiment, some other frequency decoder combinations are also tested over each database as illustrated in Table \ref{t1}. The top $n=5$ face images are retrieved and ARP is reported. It is found that the performance of FDLBP is generally better when low-pass and high-pass information are decoded except over the ESSEX database. It is also pointed out through this experiment that among the high-pass filters, the $F_{hv}$ and $F_d$ filters are more important than the $F_{sv}$ and $F_{sh}$ filters. Though, the results in previous subsection are computed for one combination of decoders (i.e., $<(F_a, F_{hv}, F_{d}), (F_a, F_{sv}, F_{sh})>$), its performance can be further boosted by opting the right decoder combination for a particular database.

\begin{table}[!t]
\caption{The ARP(\%) using the proposed FDLBP descriptor with Euclidean, Cosine, Emd, L1, D1, and Chi-square distance measures over the PaSC, LFW, PubFig, and ESSEX face databases. The number of retrieved images ($n$) is 5 in this experiment. The highest ARP values are expressed in bold for each database.}
\label{t2}
\begin{center}
\begin{tabular}{lcccc}
\hline
\\[-0.65em]\multirow{2}{*}{Distance} & \multicolumn{4}{c}{Face Databases}\\ \\[-0.65em]
\cline{2-5} 
\\[-0.65em] & PaSC & LFW & PubFig & ESSEX \\ \\[-0.65em]
\hline
\\[-0.65em]
Euclidean & 26.63 & 28.73 & 35.50 & 92.33\\ \\[-0.85em]
Cosine & 27.02 & 28.93 & 35.94 & 92.29\\ \\[-0.85em]
Emd & 23.78 & 26.09 & 32.15 & 84.27\\ \\[-0.85em]
L1 & 29.79 & 31.47 & 39.11 & \textbf{93.86}\\ \\[-0.85em]
D1 & 29.95 & 31.58 & 39.19 & \textbf{93.86}\\ \\[-0.85em]
Chi-square & \textbf{32.00} & \textbf{32.46} & \textbf{41.73} & 93.77\\ \\[-0.65em]
\hline
\end{tabular}
\end{center}
\end{table}

\subsection{Experiment with Different Similarity Measures}
The Chi-square distance is used for the similarity measures in the rest of the experiments. In this experiment, the performance of FDLBP descriptor is investigated with different distances, such as Euclidean, Cosine, Emd, L1, D1, and Chi-square. The ARP for $n=5$ number of retrieved images using the FDLBP descriptor with different similarity measures are summarized in Table \ref{t2}. It is noticed that the Chi-square distance is better suited with the FDLBP descriptor. The D1 distance is the second best performing similarity measure.

\begin{table}[!t]
\caption{The ARP(\%) using mdLBP, FDLBP, cFDLBP, and FmdLBP descriptors over the PaSC, LFW, PubFig, and ESSEX face databases. The number of retrieved images ($n$) is 5 in this experiment. The highest ARP values are expressed in bold for each database.}
\label{t3}
\begin{center}
\begin{tabular}{lcccc}
\hline
\\[-0.65em]\multirow{2}{*}{Descriptors} & \multicolumn{4}{c}{Face Databases}\\ \\[-0.65em]
\cline{2-5} 
\\[-0.65em] & PaSC & LFW & PubFig & ESSEX \\ \\[-0.65em]
\hline
\\[-0.65em]
mdLBP & 26.55 & 29.65 & 34.47 & 91.52\\ \\[-0.85em]
FDLBP & 32.00 & 32.46 & 41.73 & 93.77\\ \\[-0.85em]
cFDLBP & \textbf{32.72} & \textbf{34.46} & \textbf{42.42} & \textbf{94.46}\\ \\[-0.85em]
FmdLBP & 27.11 & 33.04 & 38.71 & 92.75\\ \\[-0.65em]
\hline
\end{tabular}
\end{center}
\end{table}

\subsection{Experiment with Color Channels}
This experiment is performed to demonstrate the performance of the frequency decoder as compared to the color decoder \cite{mdlbp}. The four descriptors are computed in the different scenarios, including 1) the original mdLBP, which is computed using the color decoder without filtering \cite{mdlbp}, 2) the FDLBP, which is computed using the frequency decoder without color consideration (i.e., over gray scale image), 3) the cFDLBP, which is computed by concatenating the frequency decoder features of each color channel, and 4) the FmdLBP, which is computed by concatenating the color decoder features of each filter. The retrieval results in terms of the ARP for $n=5$ number of retrieved images over the PaSC, LFW, PubFig, and ESSEX face databases are reported in Table \ref{t3}. The ARP of the FDLBP descriptor is greater than the ARP of the mdLBP descriptor and the ARP of the cFDLBP descriptor is greater than the ARP of the FmdLBP descriptor. Thus, it can be easily observed that the frequency decoder is far better than the color decoder over the face databases in the image retrieval framework.

\begin{table*}[!t]
\caption{The cost comparison among the different descriptors in terms of the feature computation time and storage space. The cost is computed over all the images of the dataset. The feature computation time and storage space are reported in Seconds and MB, respectively.}
\label{t4:cost}
\begin{center}
\begin{tabular}{lcccc|cccc}
\hline
\\[-0.65em]\multirow{3}{*}{Descriptors} & \multicolumn{4}{c|}{Computation Time (Seconds)} & \multicolumn{4}{c}{Storage Space (MB)}\\ \\[-0.65em]
\cline{2-9} 
\\ & \multicolumn{4}{c|}{Face Databases} & \multicolumn{4}{c}{Face Databases}\\ \\[-0.65em]
\\[-0.65em] & PaSC & LFW & PubFig & ESSEX & PaSC & LFW & PubFig & ESSEX \\ \\[-0.65em]
\hline
\\[-0.65em]
LBP & 34.97 & 6.37 & 10.71 & 41.76 & 3.8 & 1.3 & 2.8 & 3.3\\ \\[-0.85em]
LDP & 41.84 & 8.18 & 13.12 & 46.54 & 15.5 & 5.3 & 11.4 & 13.6\\ \\[-0.85em]
SOBEL\_LBP & 41.50 & 7.29 & 12.08 & 42.88 & 7.9 & 2.7 & 5.8 & 7.0\\ \\[-0.85em]
SLBP & 37.69 & 7.15 & 11.34 & 41.76 & 3.5 & 1.2 & 2.6 & 3.1\\ \\[-0.85em]
BoF\_LBP & 44.19 & 9.81 & 14.00 & 49.98 & 17.1 & 5.9 & 12.9 & 15.1\\ \\[-0.85em]
LDGP & 37.08 & 5.99 & 10.12 & 38.45 & 1.2 & 0.4 & 0.83 & 1.1\\ \\[-0.85em]
LVP & 42.76 & 7.88 & 12.47 & 45.54 & 13.5 & 4.5 & 9.8 & 12.1\\ \\[-0.85em]
SVD\_S\_LBP & 58.07 & 15.30 & 25.64 & 58.70 & 2.4 & 0.9 & 1.8 & 2.2\\ \\[-0.85em]
FDLBP & 51.91 & 13.11 & 21.26 & 60.79 & 41.0 & 13.9 & 30.4 & 35.7\\ \\[-0.65em]
\hline
\end{tabular}
\end{center}
\end{table*}

\subsection{Cost of the Descriptors}
The previous experiments clearly demonstrate the improved retrieval performance of the proposed FDLBP descriptor. The cost in terms of the feature computation time as well as the storage space is another important measure. In order to compute the cost, a desktop computer with Intel-i7 processor, 48-GB RAM, and Ubuntu16.04-64bit operating system is used. The MATLAB R2017b framework is used to run the program. The computation time in seconds and the storage space in MB for the LBP, LDP, SOBEL\_LBP, SLBP, BoF\_LBP, LDGP, LVP, SVD\_S\_LBP, and FDLBP descriptors over the PaSC, LFW, PubFig, and ESSEX face databases are reported in the Table \ref{t4:cost}. It is observed that the computation time for the proposed FDLBP descriptor is high as compared to the other descriptors except SVD\_S\_LBP descriptor. As far as storage is concerned, the proposed FDLBP descriptor requires high space as compared to the other descriptors. The high computation and storage costs are not the problem any more due to the high computational facilities and the huge amount of memory are available at reasonable price.

\section{Conclusion}
In this paper, a frequency decoder based local descriptor is proposed for the face representation. The descriptor first computes the low-pass and high-pass filtered images of the input face images. Then, it uses two decoders with three input channels each. The inputs to the decoder are computed as the binary codes with the help LBP operator over the filtered images. The feature vectors are generated over each output image of both decoders and finally combined into a single frequency decoded local binary pattern (FDLBP) feature vector. The proposed descriptor is tested in the image retrieval framework over four very challenging constrained and unconstrained face databases. The results are compared with the state-of-the-art descriptors mostly face descriptors. The experimental results reveal the superiority of the proposed descriptor. The FDLBP descriptor is also tested with different frequency decoder combinations and it is suggested that its performance can be further improved by choosing the most suitable decoder combination over a face database. It is also noticed that Chi-square distance is a better choice for the similarity measure. In case of color face images, it is suggested to use the frequency decoder over color channels instead of the color decoder over filtered images.


\ifCLASSOPTIONcaptionsoff
  \newpage
\fi

\bibliographystyle{IEEEtran}
\bibliography{ref}

\begin{IEEEbiography}[{\includegraphics[width=1in,height=1.25in,clip,keepaspectratio]{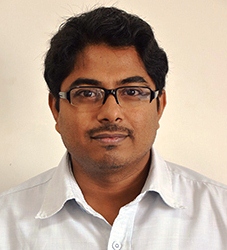}}]{Shiv Ram Dubey}
has been with the Indian Institute of Information Technology (IIIT), Sri City since June 2016, where he is currently the Assistant Professor of Computer Science and Engineering. He received the Ph.D. degree in Computer Vision and Image Processing from Indian Institute of Information Technology, Allahabad (IIIT Allahabad) in 2016. Based on his Ph.D. research work, he published 10 Journal papers including 5 IEEE Transactions/Journals. Before that, from August 2012-Feb 2013, he was a Project Officer in the Computer Science and Engineering Department at Indian Institute of Technology, Madras (IIT Madras). 

He was a recipient of several awards including the Best PhD Award in PhD Symposium, IEEE-CICT2017 at IIITM Gwalior, Early Career Research Award from SERB, Govt. of India and NVIDIA GPU Grant Award Twice from NVIDIA. He received Outstanding Certificate of Reviewing Award from Information Fusion, Elsevier in 2018 and Certificate of Reviewing Awards from Biosystems Engineering and Computers in Biology and Medicine, Elsevier in 2015 and 2016, respectively. He received the Best Paper Award in IEEE UPCON 2015, a prestigious conference of IEEE UP Section. His research interest includes Computer Vision, Deep Learning, Image Processing, Image Feature Description, Image Matching, Content Based Image Retrieval, Medical Image Analysis and Biometrics.
\end{IEEEbiography}

\end{document}